\documentclass[final]{cvpr}

\usepackage{makecell}
\usepackage{slashbox}
\usepackage{times}
\usepackage{epsfig}
\usepackage{graphicx}
\usepackage{amsmath}
\usepackage{amssymb}
\usepackage{gensymb}
\usepackage{multirow}
\usepackage{xcolor}
\usepackage{comment}
\usepackage{caption}
\usepackage{microtype}

\usepackage[pagebackref=true,breaklinks=true,letterpaper=true,colorlinks,bookmarks=false]{hyperref}
\hypersetup{
	plainpages=false,
	colorlinks=true,              %
	linkcolor=blue,               %
	anchorcolor=blue,             %
	citecolor=blue,               %
	filecolor=blue,               %
	urlcolor=blue,                %
	pdfview=FitH,                 %
	pdfstartview=FitH,            %
	pdfpagelayout=SinglePage      %
}

\def \negative_space_paragraph {-1.2em}

\def\cvprPaperID{4361} %
\def\confYear{CVPR 2021}

\title{\vspace{-1em}Single-view robot pose and joint angle estimation via render \& compare\vspace{-.5em}}

\author{
Yann Labb\'e \textsuperscript{1}
\quad
Justin Carpentier \textsuperscript{1}
\quad
Mathieu Aubry \textsuperscript{2}
\quad
Josef Sivic \textsuperscript{1,3} \\
{\textsuperscript{1} ENS/Inria \quad \textsuperscript{2} LIGM, ENPC \quad \textsuperscript{3} CIIRC CTU} \\
  {\small\url{https://www.di.ens.fr/willow/research/robopose}}
  \vspace{-.5em}
}

\begin{document}
\maketitle

\begin{abstract}
  We introduce RoboPose, a method to estimate the joint angles and the 6D camera-to-robot pose of a known articulated robot from a single RGB image. This is an important problem to grant mobile and itinerant autonomous systems the ability to interact with other robots using only visual information in non-instrumented environments, especially in the context of collaborative robotics. It is also challenging because robots have many degrees of freedom and
  an infinite space of possible configurations that often result in self-occlusions and depth ambiguities when imaged by a single camera. 
The contributions of this work are three-fold.
First, we introduce a new {\em render \& compare} approach for estimating the 6D pose and joint angles of an articulated robot that can be trained from synthetic data, generalizes to new unseen robot configurations at test time, and can be applied to a variety of robots. 
Second, we experimentally demonstrate the importance of the robot parametrization for the iterative pose updates and design a parametrization strategy that is independent of the robot structure. 
Finally, we show experimental results on existing benchmark datasets for four different robots   
and demonstrate that our method significantly outperforms the state of the art. %
Code and pre-trained models are available on the project webpage~\cite{projectpage}.

\end{abstract}

\vspace{-1.5em}

\renewcommand{\thefootnote}{\fnsymbol{footnote}}
\renewcommand*{\thefootnote}{\arabic{footnote}}
\footnotetext[1]{Inria Paris and D\'{e}partement d'informatique de l'ENS, \'{E}cole normale sup\'{e}rieure, CNRS, PSL Research University, 75005 Paris, France.}
\footnotetext[2]{LIGM, \'Ecole des Ponts, Univ Gustave Eiffel, CNRS, Marne-la-vall\'ee, France.}
\footnotetext[3]{Czech Institute of Informatics, Robotics and Cybernetics at the Czech Technical University in Prague.}

\section{Introduction}
\begin{figure}[t]
  \centering
    \vspace*{-3mm}
  \includegraphics[width=1.0\columnwidth]{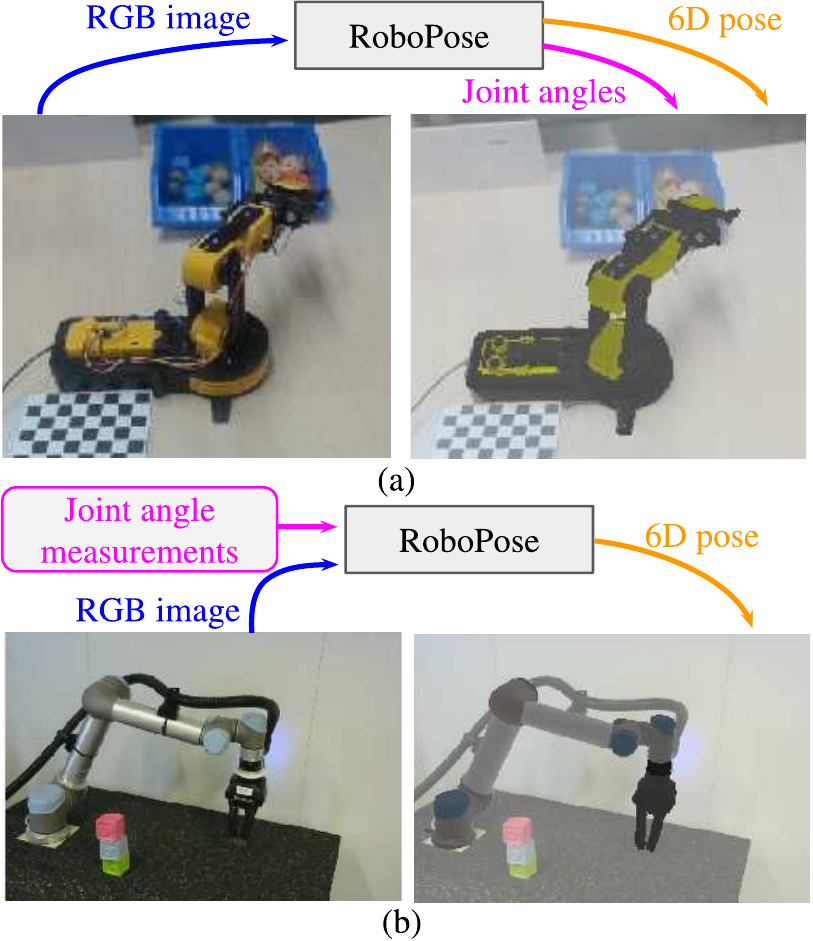}
  \small
  \vspace*{-5mm}
  \caption{\small {\bf RoboPose.} (a) Given a single RGB image of a known articulated robot in an unknown configuration (left), RoboPose estimates the joint angles and the 6D camera-to-robot pose (rigid translation and rotation) providing the complete state of the robot within the 3D scene, here illustrated by overlaying the articulated CAD model of the robot over the input image (right). 
  (b) When the joint angles are known at test-time (e.g. from internal measurements of the robot), RoboPose can use them as an additional input to estimate the 6D camera-to-robot pose to enable, for example, visually guided manipulation without fiducial markers. %
    }
  \label{fig:teaser}
\vspace*{-5mm}
\end{figure}

The goal of this work is to recover the state of a known articulated robot
within a 3D scene using a single RGB image. The robot state is defined by (i)
its 6D pose, i.e. a 3D translation and a 3D rotation with respect
to %
the camera frame, and (ii) the joint angle values of the
robot's articulations. The problem set-up is illustrated in Figure~\ref{fig:teaser}.
This is an important problem to grant mobile and itinerant autonomous systems the ability to interact with
  other robots using only visual information in non-instrumented environments. 
For instance, in the context of collaborative tasks between two or more robots, having knowledge of the pose and the joint angle values of all other robots would allow better distribution of the load between robots involved in the task~\cite{caccavale2016cooperative}.

The problem is, however, very challenging because robots can have many degrees of freedom\,(DoF) and an infinite space of admissible configurations that  often  result  in  self-occlusions  and  depth  ambiguities when imaged by a single camera. 
The current best performing methods for this problem~\cite{lee2020camera,zuo2019craves} use a deep neural network to localize in the image a fixed number of pre-defined keypoints (typically located at the articulations) and then solve a 2D-to-3D optimization problem to recover the robot 6D pose~\cite{lee2020camera} or pose and configuration~\cite{zuo2019craves}. 
For rigid objects, however, methods
based on 2D keypoints~\cite{Lowe1999-bf,bay2006surf,Collet2010-zj,Collet2011-lj,Rad2017-de,Tremblay2018-bd,Kehl2017-ek,Tekin2017-hp,peng2019pvnet,pavlakos20176,hu2019segmentation} have been recently outperformed by 
{\em render \& compare} methods that forgo explicit detection of 2D keypoints but instead use the entire shape of the object by comparing the rendered view of the 3D model to the input image and iteratively refining the object's 6D pose~\cite{Rad2017-de,zakharov2019dpod,li2018deepim,labbe2020cosypose}.
Motivated by this success, we investigate how to extend the { \em render \& compare } paradigm for articulated objects. %
This presents significant challenges. First, we need to estimate many more degrees of freedom than the sole 6D pose. Articulated robots
we consider in this work can have up to 15 degrees of freedom in addition to their 6D rigid pose in the environment. Second, the space of configurations is continuous and hence there are infinitely many configurations in which the object can appear. As a result, it is not possible to see all configurations during training and the method has to generalize to unseen configurations at test time.
Third, the choice of transformation parametrization plays an important role for 6D pose estimation of rigid objects~\cite{li2018deepim} and finding a good parametrization of pose updates for articulated objects is a key technical challenge.

\vspace{-1em}
\paragraph{Contributions.} To address these challenges, we make the following contributions.
First, we introduce a new {\em render \& compare} approach for estimating the 6D pose and joint angles of an articulated robot that can be trained from synthetic data, generalizes to new unseen robot configurations at test time, and can be applied to a large variety of robots (robotic arms, bi-manual robots, etc.). 
Second, we experimentally demonstrate the importance of the robot pose parametrization for the iterative pose updates and design an effective  parametrization strategy that is independent of the robot. 
Third, we apply the proposed method in two settings: (i) with known joint angles (e.g. provided by internal measurements from the robot such as joint encoders), only predicting the camera-to-robot 6D pose, and (ii) with unknown joint angles, predicting both the joint angles {\em and} the camera-to-robot 6D pose. We show experimental results on existing benchmark datasets for both settings that include a total of four different robots and demonstrate significant improvements compared to the state of the art.

\section{Related work}

\paragraph{6D pose estimation of rigid objects}
from RGB images \cite{Roberts1963-ck,Lowe1987-yf,Lowe1999-bf} is one of the oldest problems in computer vision. It has been successfully approached by estimating the
pose from 2D-3D correspondences obtained via local invariant features~\cite{Lowe1999-bf,bay2006surf,Collet2010-zj,Collet2011-lj}, or by
template-matching~\cite{Hinterstoisser2011-es}. Both these strategies have been
revisited using convolutional neural networks (CNNs). A set of sparse \cite{Rad2017-de,Tremblay2018-bd,Kehl2017-ek,Tekin2017-hp,peng2019pvnet,pavlakos20176,hu2019segmentation} or dense \cite{Xiang2018-dv,Park2019-od,song2020hybridpose,zakharov2019dpod} features is detected on the object in the image using a CNN and the resulting 2D-to-3D correspondences are used to recover
the camera pose using PnP~\cite{lepetit2009epnp}. The best performing
methods for 6D pose estimation from RGB images are now based on variants of the {\em render \& compare} strategy
\cite{oberweger2015training,Rad2017-de,li2018deepim,manhardt2018deep,oberweger2019generalized,labbe2020cosypose,zakharov2019dpod} and are approaching the accuracy of methods using depth as input~\cite{hodan2020bop, Hodan_undated-sl,li2018deepim,labbe2020cosypose}.

\vspace{\negative_space_paragraph}
\paragraph{Hand-eye calibration} (HEC)~\cite{Horaud1995-fl,Heller2011-so}
methods recover the
6D pose of the camera with respect to a robot. The most common approach is to detect in the image fiducial markers~\cite{garrido2014automatic,fiala2005artag,olson2011apriltag} placed on the
robot at known positions. The resulting 3D-to-2D correspondences are then used to recover the camera-to-robot pose
using {\it known} joint angles and the kinematic description of the robot by
solving an optimization problem~\cite{park1994robot,ilonen2011robust,yang1994calibrating}.
Recent works have explored using CNNs~\cite{lambrecht2019towards,lee2020camera} to perform this task by recognizing 2D keypoints at specific robot parts and using the resulting 3D-to-2D correspondences to recover the hand-eye calibration via PnP. The most recent work in this direction~\cite{lee2020camera} 
demonstrated that such learning-based approach could replace
more standard hand-eye calibration methods~\cite{tsai1989new} to perform online calibration and object manipulation~\cite{tremblay2020indirect}. Our {\em render \& compare} method significantly outperforms~\cite{lee2020camera} 
and we also demonstrate that our method can achieve a competitive accuracy without requiring known joint angles at test time. %

\vspace{\negative_space_paragraph}
\paragraph{Depth-based pose  estimation  of  articulated  objects.} Previous work on this problem
can be split into three classes. %
The first class of methods aims at discovering properties of
the kinematic chain through active manipulation~\cite{katz2008manipulating,katz2013interactive,hausman2015active,martin2016integrated}  using depth as input and unlike our approach cannot be applied to a single image. The second class of methods aims at recovering all parameters of the kinematic chain from a single RGBD image, including the joint angles, without knowing the specific articulated object~\cite{li2020category,yi2018deep,abbatematteo2019learning,zhou2016simultaneous}. In contrast, we focus on the set-up with a known 3D model, e.g. a specific type of a robot.
The third class of methods, which is closest to our set-up, considers pose and joint angle estimation \cite{michel2015pose,desingh2019factored,pauwels2014real} for known articulated objects but relies on depth as input and only considers relatively simple kinematic chains such as laptops or drawers where the joint
parameters only affect the pose of one part. Others recover joint
angles of a known articulated robot part~\cite{bohg2014robot,widmaier2016robot}
but do not recover the 6D pose of the camera and also rely on depth.
In contrast, our method accurately estimates the
pose and joint angles of a robotic arm with many degrees of freedom from a single RGB image.

\vspace{\negative_space_paragraph}
\paragraph{Robot pose and joint angle estimation from an RGB image.} 
To the best of our knowledge,
only~\cite{zuo2019craves} has addressed a scenario similar to us where the robot pose and joint angles are estimated together from a single RGB image.  
A set of predefined 2D keypoints is recognized in the image and the 6D pose and joint angles are then recovered by solving a nonlinear non-convex optimization problem. Results are shown on a 4 DoF robotic arm. In contrast, we describe a new {\em render \& compare} approach for this problem, demonstrate significantly improvements in 3D accuracy and show results on robots with up to 15 DoF.

\section{Approach}

\label{sec:method}
We present our {\em render \& compare} framework to
recover the state of a robot within a 3D scene given a single RGB image. We assume the camera intrinsic parameters, the CAD model and kinematic description of the robot are known. We start by formalizing the problem in Section~\ref{sec:problem_definition}. 
We then present an overview of our approach in Section~\ref{sec:framework_overview} and explain our training in Section~\ref{sec:training}. 
Finally, we detail the key choices in the problem parametrization in Section~\ref{sec:choice_of_coordinate_system}. %

\subsection{Problem formalization}
\label{sec:problem_definition}

\begin{figure}[t]
  \centering
  \includegraphics[width=1.0\columnwidth]{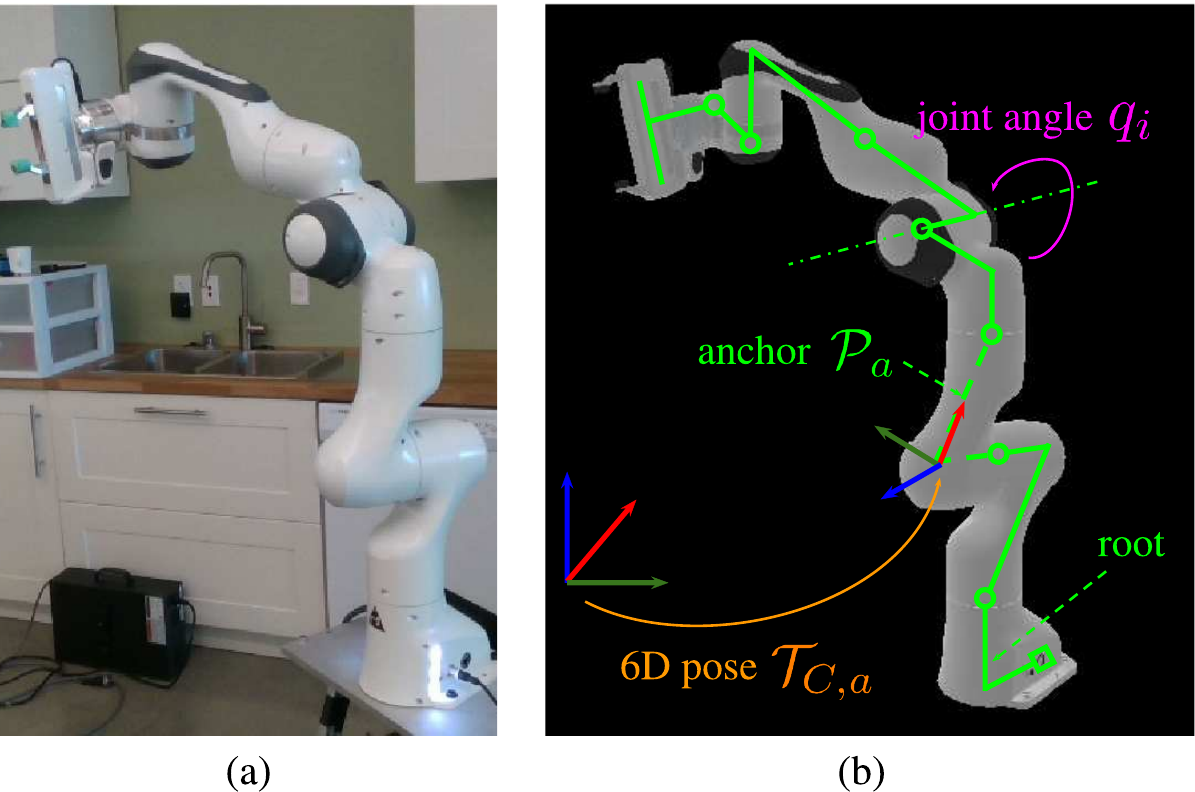}
  \caption{\small \textbf{Problem definition.} Given an RGB image (a) of a known robot, the goal
    is to recover (b) the 6D pose $\mathcal{T}_{C,a}$ of an anchor part $\mathcal{P}_a$ with respect to the camera frame and all the
    joint angles $q_i$ of the known robot kinematic description (in green).}
  \label{fig:problem_definition}
  \vspace{-1.2em}
\end{figure}

Our notations are summarized in Figure~\ref{fig:problem_definition}.
We consider a known robot composed of \textit{rigid parts} $\mathcal{P}_0$,...,$\mathcal{P}_N$ whose 3D models are known.  An articulation, or \textit{joint}, connects a parent part to a child part. Given the \textit{joint angle} $q_i$ of the $i$-th joint, we can retrieve the relative 6D transformation between the parent 
and child reference frames. %
Note that for simplicity we only consider revolute joints, i.e. joints parametrized by a single scalar angle of
rotation $q_i$, but our method is not specific to this type of joints. 
The rigid parts and the joints define the \textit{kinematic tree} of the robot.
This kinematic description can be used to compute the relative 6D pose between any two parts of the robot.
In robotics, the full state of a robot $\mathcal{S}$ is defined by the joint angles and the 6D pose of
the {\it root} of the kinematic tree.
Defining the 6D pose of the robot with respect to the root (whose pose is independent of the joint angles since it is not a child of any joint) is a crucial choice in the parametrization of the problem, but also arbitrary, since an equivalent kinematic tree could be defined using any part as the root. We instead define the full state of the robot  %
by (i) the selection of an {\it anchor part} $\mathcal{P}_a$, (ii) the 6D pose of the anchor
with respect to the camera $\mathcal{T}_{C,a}$, and (iii) the joint
angles $q=(q_1, ..., q_D)\in\mathbb{R}^{D}$, where $D$ is the number of joints. %
Note the anchor part can change across iterations of our approach.
We discuss the choice of the anchor in Section~\ref{sec:choice_of_anchor} and experimentally demonstrate it has an important influence on the results. %

\subsection{Render \& compare for robot state estimation}
\label{sec:framework_overview}

\begin{figure*}[t]
  \centering
  \includegraphics[width=1.0\linewidth]{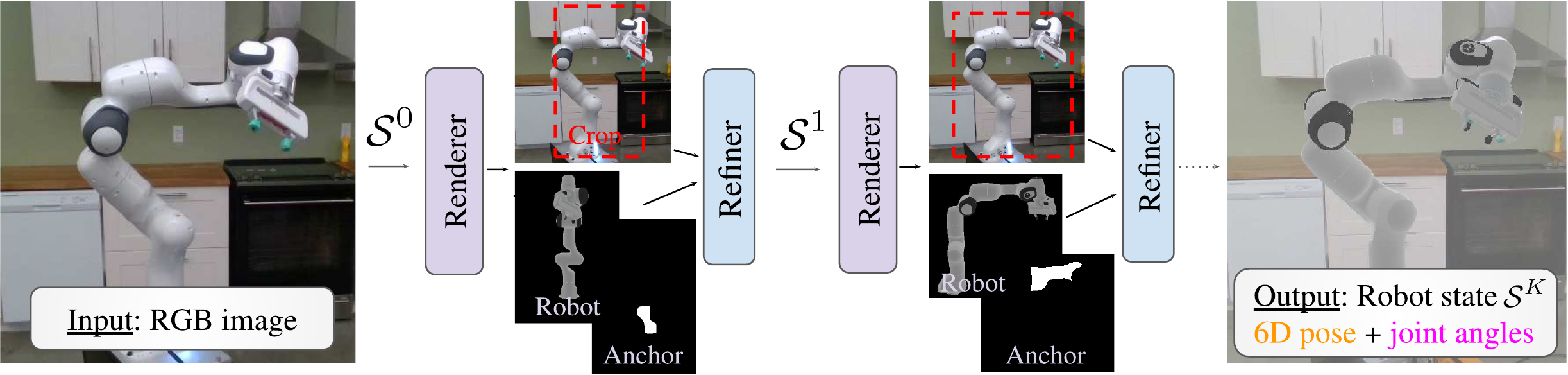}
  \vspace*{-1.8em}
  \caption{\small {\bf RoboPose overview.} Given a single input RGB image,
 the state $S$ (6D camera-to-robot pose and joint angles) of the robot is iteratively updated using renderer and refiner modules to match the input image.  The refinement module takes as input the cropped observed image and a rendering of the robot as well as the mask of an anchor part. The anchor part is used for updating the rigid 6D pose of the robot while the rest of the parts are updated by changing their joint angles. Note that the anchor part is changing across iterations making the refinement more robust.}
  \label{fig:render_compare_overview}
  \vspace{-1.3em}
\end{figure*}

We now present our iterative deep {\em render \& compare} framework, %
illustrated in Figure~\ref{fig:render_compare_overview}. %
We iteratively refine the state estimate as follows. First, given a current estimate of the state $\mathcal{S}^k$ we render an RGB image of the robot $\mathcal{R}(S^{k})$ and the mask of the anchor part. %
We then apply a deep refiner network that takes as input crops of the rendered image and %
the input RGB image $I$ of the scene. It outputs a new state of the robot $\mathcal{S}^{k+1}=f_\theta(\mathcal{S}^k, I)$ to attempt to match the ground truth state $\mathcal{S}^{gt}$ of the observed robot. Unlike prior works that have used
{\em render \& compare} strategies for estimating the 6D pose of rigid objects
\cite{li2018deepim,zakharov2019dpod,labbe2020cosypose}, our method does not
require a coarse pose estimate as initialization.

\vspace{-1.0em}
\paragraph{Image rendering and cropping.} To render the image of the robot we use a fixed focal length (defining an intrinsic camera matrix) during training. The rendering is fully defined by the state of the robot and the camera matrix. Instead of giving to the refiner network the full image and the rendered view, we focus the inputs on the robot by cropping the images as follows. We project the centroid of the rendered robot in the image, consider the smallest bounding box of aspect ratio 4/3 centered on this point that encloses the projected robot and increase its size by 40\% (see details in the appendix). %
{This crop depends on the projection of the robot to the input image that varies during training, and thus provides an augmentation of the effective focal length of the virtual cropped cameras. Hence, our method can be applied to cameras with different intrinsics at test time as we show in our experiments.}

\vspace{\negative_space_paragraph}
\paragraph{Initialization.} We initialize the robot to a state
$\mathcal{S}^0$ defined by the joint configuration $q^0$ and the pose $\mathcal{T}_{C,a}^0$ of the anchor part $a$ with respect to the camera $C$. At training time we define $\mathcal{S}^0$ using perturbations of the ground truth state. At test time we initialize %
the joints to the middle of the
joint limits, %
and the initial pose $\mathcal{T}_{C,a}^0$ so that the frame of the robot base is aligned with the camera frame %
and the 2D bounding box defined by the projection of
the robot model approximately matches %
the size of image. More details are given in the appendix.

\vspace{-1em}
\paragraph{Refiner and state update.}
At iteration $k$, the refiner predicts an update $\Delta q^k$ of the joint angles $q^k$ (composed of one scalar angle per joint), such that \begin{equation}
  q^{k+1} = q^{k} + \Delta q^k,
\end{equation}
and an update $\Delta \mathcal{T}^k$ of the current 6D pose $\mathcal{T}^k_{C,a}$ of the anchor part, such that
\vspace{-3mm}
\begin{equation}
  \label{eq:pose_update_rule}
  \mathcal{T}_{C,a}^{k+1}=\mathcal{T}^k_{C,a} \circ \Delta \mathcal{T}^k,
\end{equation}
where we follow DeepIM~\cite{li2018deepim}'s parametrization for pose update $\Delta
\mathcal{T}^k$. This parametrization disentangles rotation and translation prediction but
crucially depends on the choice of a \textit{reference point} we
call $O$. In DeepIM this point is simply taken
as the center of the reference frame of the rigid object but there is not such a
natural choice of reference point for articulated objects, which have
multiple moving parts.  We discuss several possible choices of the reference point $O$ in Sec.~\ref{sec:choice_of_coordinate_system} and demonstrate experimentally  it has an important impact on the results. In particular, we show that naively selecting the reference frame of the root part is sub-optimal.

\subsection{Training}
\label{sec:training}

In the following, we describe our loss function, %
synthetic training data, %
implementation details and discuss how to best use known joint angles if available. %

\vspace{\negative_space_paragraph}
\paragraph{Loss function.}
We train our refiner network using the following loss:
\begin{equation}
  \label{eq:loss_general}
  \hspace{-1em}
  \mathcal{L}(\theta) = \sum_{k=0}^{K-1} \mathcal{L}_{a}(\mathcal{T}^{k}_{C,a},\Delta \mathcal{T}^{k}, \mathcal{T}_{C,a}^{gt})\mathcal{} + \lambda\, \mathcal{L}_{q}(q^{k},\Delta q^k, q^{\mathrm{gt}}), 
\end{equation}
where $\theta$ are the parameters of the refiner network, $K$ is the maximum number of iterations of the refinement algorithm, $\mathcal{T}_{C,a}^{gt}$ is the ground truth 6D pose of the anchor, $q^{\mathrm{gt}}$ are the ground truth joint angles and $\lambda$ is a hyper-parameter to balance between the 6D pose loss $\mathcal{L}_a$ and the joint angle loss $\mathcal{L}_q$. 
The 6D pose loss $\mathcal{L}_a$ measures the distance between the predicted 3D point cloud obtained using $\mathcal{T}^{k}_{C,a}$ transformed with $\Delta
\mathcal{T}^{k}$ and the ground truth 3D point cloud (obtained using $\mathcal{T}_{C,a}^{gt}$) of the {anchor} $\mathcal{P}_a$. We use the same loss as~\cite{labbe2020cosypose} that disentangles rotation, depth and image-plane translations~\cite{Simonelli2019-da} (see equations in the appendix).
For $\mathcal{L}_q$, we use a simple $L_2$ regression loss, $\mathcal{L}_{q} = \|q^{k}+\Delta q^k - q^{gt}\|_2^2$.  

Note that the 6D pose loss is measured only on the anchor part $a$ while the alignment of the other parts of the robot is measured by the error on their joint angles (rather than alignment of their 3D point clouds). This disentangles the 6D pose loss $\mathcal{L}_a$ from the joint angle loss $\mathcal{L}_q$ and we found this leads to better convergence. %
We sum the loss over the refinement iterations $k$ to imitate how the refinement algorithm is applied at test time but the error gradients are not backpropagated through rendering and iterations.
Finally, for simplicity the loss~\eqref{eq:loss_general} is written for a single training example, but we sum it over all examples in the training set. %

\vspace{\negative_space_paragraph}
\paragraph{Training data.} 
For training the refiner, we use existing datasets~\cite{lee2020camera, zuo2019craves}
provided by prior works for the Kuka, Panda, Baxter, OWI-535 robots.
All of these datasets are synthetic, generated using similar procedures based on
domain randomization~\cite{tobin2017domain,Loing2018,Sadeghi2017-dr,James2018-db}. The joint angles are sampled independently and uniformly within their
bounds, without assuming further knowledge about their test time distribution.
We add data augmentation similar to~\cite{labbe2020cosypose}. %

We sample the initial state $\mathcal{S}^0$ by adding noise
to the ground truth state in order to simulate errors of the network prediction at the previous state of the refinement as well as the error at the initalization. %
For the pose, we sample a translation from a centered Gaussian
distribution with standard deviation of $10$\,cm, and a rotation by sampling three angles from a centered Gaussian distribution of standard deviation 
$60^\circ$.
For the joint angles, we sample an additive noise from a centered Gaussian distribution with a standard deviation equal to $5\%$ of the
joint range of motion, which is around $20^{\circ}$ for most of the joints of the robots we considered.

\vspace{\negative_space_paragraph}
\paragraph{Implementation details.}

We train separate networks for each robot. We use a standard ResNet-34 architecture~\cite{he2016deep} as the backbone of the deep refiner.  The hyper-parameters are $\lambda=1$ and $K=3$ training iterations. Note that at test time we can perform more iterations, and the results we report correspond to 10 iterations. The anchor is sampled randomly among the 5 largest parts of the robot at each iteration. This choice is motivated in Section~\ref{sec:parametrization_choices} and other choices are considered in the experiments, Section~\ref{sec:exp_ablations}.
We initialize the network parameters randomly and perform the optimization using Adam \cite{Kingma2014-xz}, with the procedure described in the appendix for all the networks.

\vspace{\negative_space_paragraph}
\paragraph{Known joint angles at test time.} The approach described previously could be used at test time with measured joint angles $q^0=q^{gt}$ and by ignoring the joint update, but we observed better results by training a separate network which only predicts a pose update for this scenario.  In this context where the joint values are known and constant, the full robot is considered as a single and unique anchor. Yet, the problem remains different from classic rigid object 6D object pose estimation because the network must generalize to new joint configurations unseen during training.

\subsection{Parametrization choices}
\label{sec:parametrization_choices}

There are two main parametrization choices in our approach: (i) the choice of the reference point $O$ for the parametrization of the pose update $\Delta \mathcal{T}^k$ in equation~\eqref{eq:pose_update_rule} and (ii) the choice of the anchor part to update the 6D pose and measure pose loss in equation~\eqref{eq:loss_general}. These choices have a significant impact on the results, as shown in Section~\ref{sec:experiments}.

\vspace{\negative_space_paragraph}
\paragraph{Choice of the reference point for the pose update.}
\label{sec:choice_of_coordinate_system}
Similar to~\cite{li2018deepim}, we parametrize the pose update as a rotation around a reference point $O$ and a translation defined as a function of the position of $O$ with respect to the camera. The fact that the rotation is around $O$ is a first obvious influence of this choice on the transformation that needs to be predicted.
The impact on the translation update parameters is more complicated: they are defined by a multiplicative update on the depth of $O$ and by an equivalent update in pixels in the image, which is also related to the real update by the depth of $O$ (see equations in the appendix).

A seemingly natural choice for reference point $O$ would be a physical point on the robot, for example the center of the base of the robot or the anchor part. %
However, on the contrary to the rigid object case, if that part is not visible or is partially occluded, the network cannot infer the position of the reference $O$ precisely, and thus cannot predict a relevant translation and rotation update. In experiments, we show it is better to use as $O$ the centroid of the estimated robot state, which takes into account the estimated joint configuration, and can be more reliably estimated.

\begin{figure}[t]
  \centering
  \includegraphics[width=1.0\linewidth]{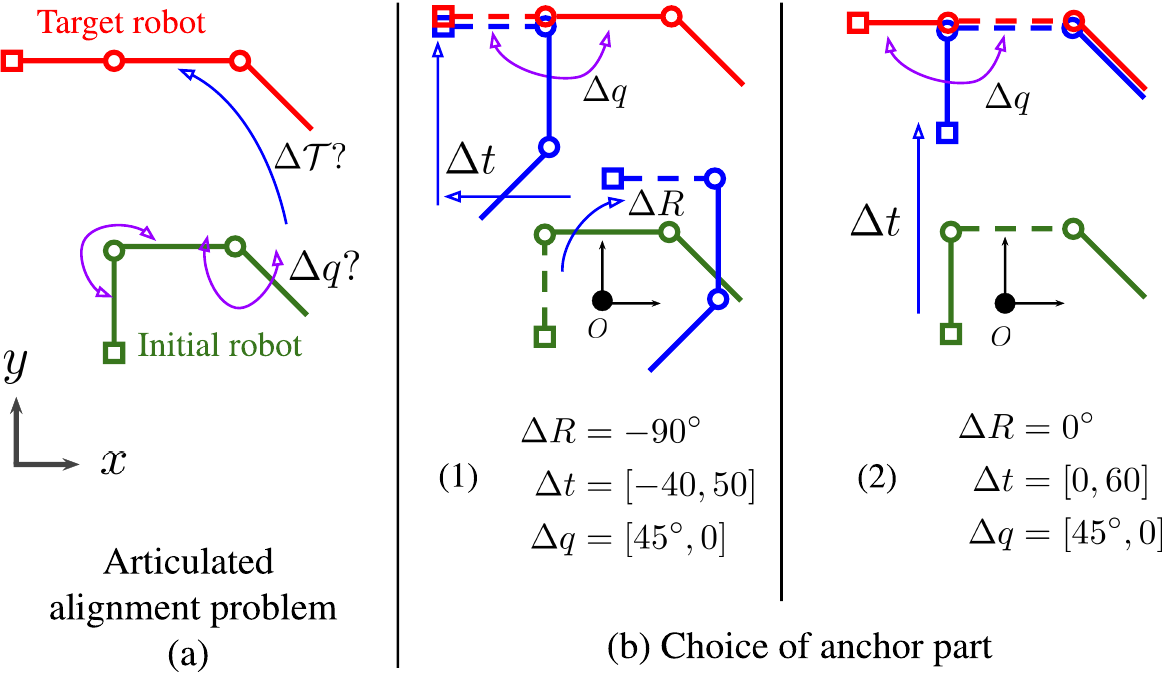}
  \caption{\small {\bf Choice of the anchor part.}
    {{\bf (a)} We analyze
    how the choice of the anchor part affects 
    the complexity of the rigid alignment  $\Delta \mathcal{T}$ and the joint angle update $\Delta q$ to align an initial state of the robot (green) with the target state of the robot (red). {\bf (b)} We show the required rigid pose update (composed of a rotation and a translation) and the required joint update for two different choices of the anchor part (shown using a dashed line). In (1), the required pose update of the anchor part consists of successively applying rotation $\Delta R$ and translation $\Delta t$ along $x$ and $y$ axes (in blue). In (2), the anchor part is aligned using only a translation along the $y$ axis resulting in a simpler solution compared to (1). These examples illustrate that the choice of the anchor can have a significant impact on the complexity of the alignment problem.}
  }
  \label{fig:pose_update_unknown_joints}
  \vspace{-1.3em}
\end{figure}

\vspace{\negative_space_paragraph}
\paragraph{Choice of the anchor part.}
\label{sec:choice_of_anchor}
The impact of the choice of the anchor part $\mathcal{P}_a$ used for computing the 6D pose loss in equation~\eqref{eq:loss_general}, is illustrated in Figure~\ref{fig:pose_update_unknown_joints}.
We explore several choices of anchor part in our experiments, and show that this choice has a significant impact on the results. Since the optimal choice depends on the robot, and the observed pose, we introduce a strategy where we  randomly select the anchor among the largest parts of the robot, during both training and refinement, and show that on average it performs similarly or slightly better than the optimal oracle choice of a single unique anchor on the test set. %

\vspace{-.3em}
\section{Experiments}
\setlength{\tabcolsep}{1pt}
\begin{table*}[t]
  \small
  \centering
    \begin{tabular}{l|ccccc|cccc||c|}
      \cline{2-11}
      \multicolumn{1}{l|}{}& \multicolumn{5}{c|}{Dataset informations}& DREAM \cite{lee2020camera}          & DREAM \cite{lee2020camera} & DREAM \cite{lee2020camera}  & Ours       & \textcolor{blue}{Ours} \\
      Robot & Robot (DoF)  & Real        & \# images     & \# 6D poses  & cam. & VGG19-F   & VGG19-Q  & ResNet101-H & ResNet34 & \textcolor{blue}{Unknown angles}  \\ 
      \hline
      \multicolumn{1}{|l|}{Baxter DR}  & Baxter (15) & $\times$  & 5982       & 5982  & GL & - & 75.47 & -     & \textbf{86.59} & \textcolor{blue}{\textbf{32.66}} \\
      \hline
      \multicolumn{1}{|l|}{Kuka DR}    & Kuka (7) & $\times$   & 5997       & 5997 & GL   & - & -     & 73.30 & \textbf{89.62} & \textcolor{blue}{\textbf{80.17}}  \\
      \multicolumn{1}{|l|}{Kuka Photo}  & Kuka (7) & $\times$   & 5999       & 5999  & GL & -   & -  & 72.14  & \textbf{86.87}  & \textcolor{blue}{\textbf{73.23}}  \\
      \hline
      \multicolumn{1}{|l|}{Panda DR}       & Panda (8) & $\times$   & 5998        & 5998  & GL & 81.33   & 77.82  & 82.89     & \textbf{92.70}  & \textcolor{blue}{\textbf{82.85}}  \\
      \multicolumn{1}{|l|}{Panda Photo}    & Panda (8) & $\times$   & 5997    & 5997  & GL &  79.53   & 74.30  & 81.09  & \textbf{89.89}  & \textcolor{blue}{\textbf{79.72}}  \\
      \multicolumn{1}{|l|}{Panda 3CAM-AK}  & Panda (8) & \checkmark   & 6394  & 1  & AK &  68.91   & 52.38  & 60.52  & \textbf{76.54}  & \textcolor{blue}{\textbf{70.37}}  \\
      \multicolumn{1}{|l|}{Panda 3CAM-XK}  & Panda (8) & \checkmark   & 4966  & 1  & XK &  24.36  & 37.47  & 64.01  & \textbf{85.97}  & \textcolor{blue}{\textbf{77.61}}  \\
      \multicolumn{1}{|l|}{Panda 3CAM-RS}  & Panda (8) & \checkmark   & 5944  & 1  & RS & 76.13  & 77.98  & \textbf{78.83}  & 76.90  & \textcolor{blue}{\textbf{74.31}}  \\
      \multicolumn{1}{|l|}{Panda ORB}  & Panda (8) & \checkmark   & 32315     & 27 & RS &  61.93  & 57.09  & 69.05 & \textbf{80.54}  & \textcolor{blue}{\textbf{70.39}}  \\
      \hline
    \end{tabular}
  \caption{\small Comparison of RoboPose (ours) with the state-of-the-art approach
    DREAM~\cite{lee2020camera} for the camera-to-robot 6D pose estimation task
    using the 3D reconstruction ADD metric (higher is better). The robot joint configuration is assumed to be known (results in black) and is different in each of the image in the dataset, but the pose of the camera with respect to the robot can be fixed (\# number of 6D poses). Multiple cameras are considered to capture the input RGB images: synthetic rendering (GL), and real Microsoft Azure (AK), Microsoft Kinect360 (XK)
    and Intel RealSense (RS), which all have different intrinsic parameters. Our results in blue do not use ground truth joint angles (see Section~\ref{sec:exp_unknown_joint_angles})
    and the accuracy of the robot 3D reconstruction is evaluated using both the
    estimated 6D pose and the joint angles.}
  \label{tab:comparison_dream}
  \vspace{-1.3em}
\end{table*}

\label{sec:experiments}
We evaluate our method on
recent benchmarks for the following two tasks: (i) camera-to-robot 6D pose estimation for three widely used manipulators (Kuka iiwa7, Rethink robotic Baxter, Franka Emika
Panda)~\cite{lee2020camera}, and (ii) full state estimation of the low-cost 4\,DoF robotic arm OWI-535~\cite{zuo2019craves}.
In Section~\ref{sec:exp_known_joint_angles}, we consider the first task,
where an image of a robot with fixed known joint angles is used to estimate the 6D camera-to-robot pose. We show that our approach outperforms the
state-of-the-art DREAM method~\cite{lee2020camera}. In
Section~\ref{sec:exp_unknown_joint_angles}, we evaluate our full approach where both the 6D pose and joint angles are unknown. We show our method outperforms the state-of-the-art method~\cite{zuo2019craves} for this problem on their dataset depicting the low-cost 4\,DoF robot and that it can recover the 6D pose {\em and} joint angles
of more complex robotic manipulators. %
Finally, Section~\ref{sec:exp_ablations}
analyzes the parametrization choices discussed in Section~\ref{sec:parametrization_choices}.

\subsection{6D pose estimation with known joint angles}
\label{sec:exp_known_joint_angles}

\paragraph{Datasets and metrics.} We focus on the %
datasets annotated with 6D pose and joint angle measurements
recently introduced by the state-of-the-art method for single-view
camera-to-robot calibration, DREAM~\cite{lee2020camera}. We use the provided training datasets with 100k images generated with domain
randomization. Test splits are available as well as photorealistic
synthetic test images (Photo). For the Panda robot, real datasets are
also available. The Panda 3CAM datasets display the fixed robot performing various
motions captured by 3 fixed different cameras with different focal lengths and
resolution, all of which are different than the focal length used during
training. The largest dataset with the more varied viewpoints is Panda-ORB
with 32,315 real images in a kitchen environnement captured from 27 viewpoints with
different joint angles in each image.

We use the 3D reconstruction ADD metric which directly measures the
pose estimation accuracy, comparing distances between 3D keypoints defined at
joint locations of the robot in the ground truth and predicted pose. We refer
to the appendix for exact details on the evaluation protocol of our comparison with DREAM~\cite{lee2020camera}.

\vspace{\negative_space_paragraph}
\paragraph{Comparison with DREAM \cite{lee2020camera}.}
We train one network for each robot using the same synthetic datasets as~\cite{lee2020camera}
and report our results in Table~\ref{tab:comparison_dream}.
Our method achieves significant improvements across datasets and robots except on
Panda 3CAM-RS where the performance of~\cite{lee2020camera} with ResNet101-H variant is similar to ours. On the Panda 3CAM-AK and Panda 3CAM-XK datasets, the performance of our method is significantly higher than the ResNet101-H model of ~\cite{lee2020camera} (e.g. $+21.96$ on
3CAM-XK), which suggests that the approach of~\cite{lee2020camera} based on 2D keypoints is more
sensitive to some viewpoints or camera parameters.
Note that our method trained with the synthetic GL camera can be applied to different real cameras with different intrinsics at test time thanks to our cropping strategy which provides an augmentation of the effective focal length during training.

On Panda-ORB, the largest
real dataset that covers multiple camera viewpoints, our method achieves a
large improvement of $11.5$ points. Our performance on the synthetic datasets for
the Kuka and Baxter robots is also significantly higher
than~\cite{lee2020camera}. We believe the main reason for this large improvements is the fact that our Render \& Compare approach can directly use the shape of
the entire robot rendered in the observed configuration %
for estimating the pose rather than detecting a small number of keypoints.
\begin{figure*}[t]
  \centering
  \includegraphics[width=1.0\linewidth]{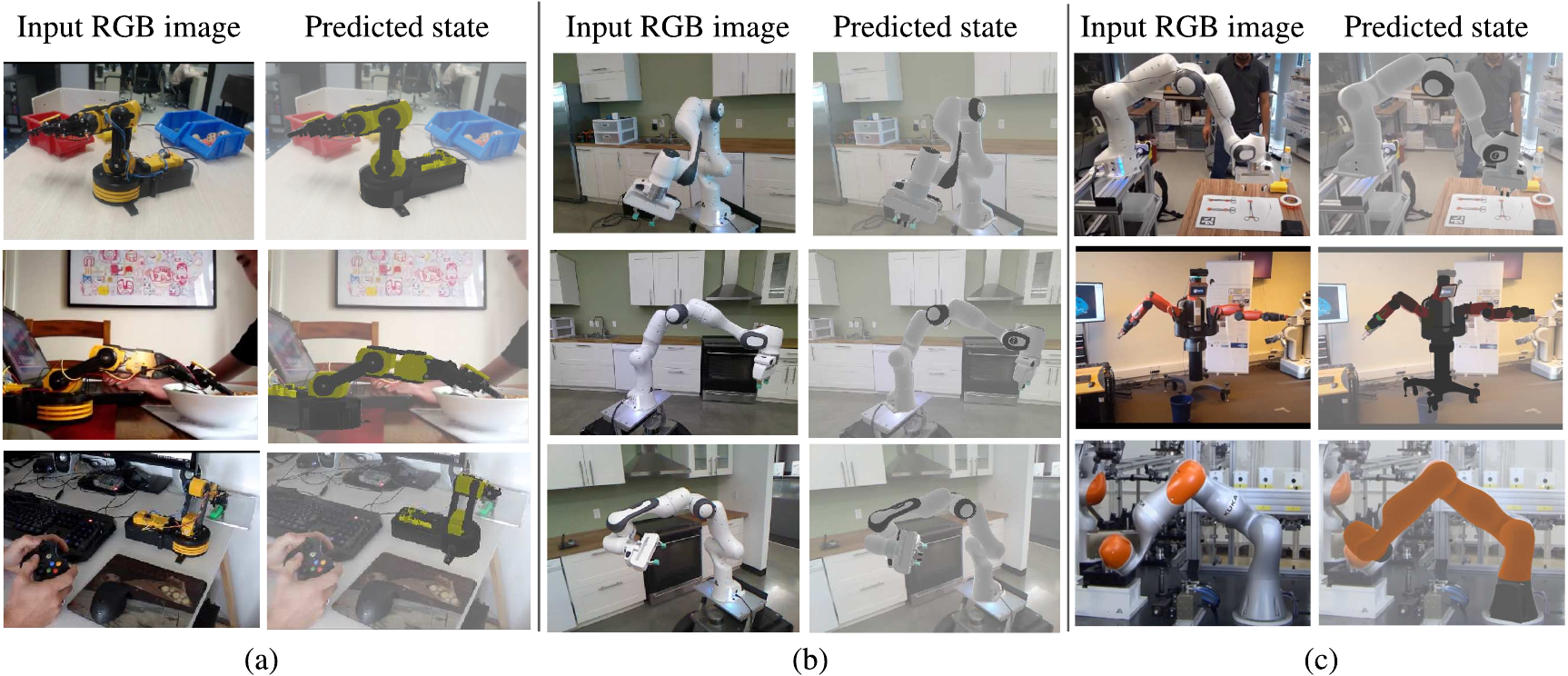}
      \vspace*{-2.0em}
  \caption{\small Qualitative results of RoboPose 6D pose and joint angle estimation for four different robots. (a) The OWI-535 robot from the CRAVES-lab (first row) and CRAVES-youtube (second and third row) datasets, (b) the Panda robot from the Panda 3CAM dataset and (c) the Panda, Baxter and Kuka robots on example images from the Internet. {\bf Please see additional results in the appendix and in the supplementary video on the project webpage~\cite{projectpage}.}} \label{fig:qualitative_results}
      \vspace*{-0.8em}
\end{figure*}

\vspace{-1.0em}

\setlength{\tabcolsep}{3pt}
\begin{table}[t]
  \small
  \centering
  \begin{tabular}{lccccc|c|c}
    \cline{2-8}
    & \multicolumn{5}{c|}{Ours (individual frames)} & Ours (online) & DREAM \cite{lee2020camera} \\
    & K=1  & K=2  & K=3  & K=5  & K=10  & K=1 & ResNet101-H \\
    \hline
    ADD & 28.5 & 72.8 & 79.1 & 80.4 & \textbf{80.7} & 80.6  & 69.1 \\
    \hline
    FPS & 16  & 8 & 4 & 2 & 1  & 16 & \textbf{30}  \\
    \hline
  \end{tabular} \caption{\small Benefits of iterative refinement and running time on Panda-ORB video sequence of robot trajectories. We report ADD and running time (frames per second, FPS) for a varying number of refiner iterations $K$. The frames are either considered individually, or the estimate is used to initialize the refiner in the subsequent frames (online) without additionnal temporal filtering.}
  \label{tab:running_time}
  \vspace{-1.2em}
\end{table}

\vspace{-0.3em}
\paragraph{Running time.} In Table~\ref{tab:running_time} we report the
running time of our method on the Panda-ORB dataset which consists of robot motion videos captured from 27 different viewpoints.
The first observation is that the accuracy increases with the number of refinement iterations $K$ used at test-time, and the most significant improvements are during the first 3 iterations. The importance of using multiple network iterations during training is further discussed in the appendix. We also report an online version of our approach that leverages temporal continuity. It runs the refiner with $K=10$ iterations on the first frame of the video and then uses the output pose as the initialization for the next frame and so on,
without any additional temporal filtering of the resulting 6D poses. This version runs at 16 frames per second (FPS) and achieves a similar performance as the full approach that considers each frame independently and runs at 1 FPS.

\subsection{6D pose and joint angle estimation}

\label{sec:exp_unknown_joint_angles}
We now evaluate the performance of our method in the more challenging scenario where the
robot joint angles are unknown and need to be estimated jointly with the 6D pose from a single RGB image.  Qualitative results on the considered datasets  as well as on real images crawled from the web are shown in Figure~\ref{fig:qualitative_results}. {Please see the appendix for additional qualitative examples, and the project page~\cite{projectpage} for a movie showing our predictions on several videos.}

\begin{table}[t]
  \newcommand{\mc}[1]{\multicolumn{2}{c}{#1}}
  \small
  \begin{center}
    \begin{tabular}{l|cc|cc|cc}
      \hline
      & \multicolumn{2}{c}{CRAVES \cite{zuo2019craves}} & \multicolumn{2}{c}{CRAVES \cite{zuo2019craves}} & \multicolumn{2}{c}{ours}\\
      & \multicolumn{2}{c}{synt} & \multicolumn{2}{c}{synt+real*} & \multicolumn{2}{c}{synt} \\
      \hline
      PCK@0.2 & \mc{95.66}    & \mc{\textbf{99.55}} & \mc{99.20} \\
      \hline
      \hline
      Error &  &  & all & top 50\% & all & top 50\% \\
      \hline
      Joints (degrees) &     &   & 11.3 & 4.74  & \textbf{5.49} & \textbf{3.22} \\
      Trans xyz. (cm)  &     &   &10.1 & 5.52  & \textbf{0.61} & \textbf{0.42} \\
      Trans norm. (cm) &     &   &19.6 & 10.5  & \textbf{1.31} & \textbf{0.90} \\
      Rot.   (degrees) &     &   &10.3 & 5.29  & \textbf{4.12} & \textbf{2.91} \\
      \hline
    \end{tabular}
  \end{center}
  \vspace{-1.5em}
  \caption{\small Results on the CRAVES-lab \cite{zuo2019craves} dataset with unknown joint angles. We report average errors on \textit{all} the images of the dataset, or on the \textit{top 50\%} images selected according to the best joint angle accuracy with respect to the ground truth. Networks are trained on synthetic data only (synt) or also using non-annotated real images of the robot (synt+real*).}
  \label{tab:craves_lab}
  \vspace{-1.6em}
\end{table}

\vspace{\negative_space_paragraph}
\paragraph{Comparison with CRAVES~\cite{zuo2019craves}.} CRAVES~\cite{zuo2019craves} is the state-of-the-art approach for this task.  We consider the two datasets used in~\cite{zuo2019craves}. {CRAVES-lab} displays the OWI-535 4DoF in a lab environment and contains 20,000 RGB images of
which 428 key frames are annotated with 2D robot keypoints, ground truth joint angles
(not used by our method) and camera intrinsics. {CRAVES-youtube} is the
second dataset containing real-world images crawled from YouTube {depicting large variations in viewpoint, illumination conditions and robot appearance. It contains 275 frames annotated with 2D keypoints but no camera intrinsic parameters, 6D pose or joint angle ground truth.}
In addition to metrics that measure the 6D pose and joint angle estimates, we report a 2D keypoint metric, PCK (percentage of keypoints), following \cite{zuo2019craves}. We refer to the appendix for details of the metrics and the evaluation protocol.

We compare with two variants of CRAVES, one trained only on synthetic images (synt),
and one that also requires real non-annotated images (synt+real*). Our method is trained only using the $5,000$ provided synthetic images. We report results on CRAVES-lab in Table~\ref{tab:craves_lab}. 
To compare with the 2D keypoint metric PCK@0.2, we project in the image the 3D keypoints of our estimated robot state.  
 On this metric, our method outperforms CRAVES trained only on synthetic images and achieves a near-perfect score, similar to their approach trained with real images. More importantly, we achieve much better results when comparing with the 3D metrics (joint angles error and translation/rotation error). CRAVES achieves high average errors when all images of the datasets are considered, which is due to the complexity of solving the nonlinear nonconvex 2D-to-3D optimization problem for recovering 6D pose and joint angles given 2D keypoint positions. %
Our method trained to directly predict the 6D pose and joint angles, achieves large improvements in precision. We reduce the translation error by a factor of 10, demonstrating robustness to depth ambiguities.

\setlength{\tabcolsep}{2pt}
\begin{table}[t]
  \small
  \centering
  \begin{tabular}{lcc|cccc||c}
    \cline{2-8}
    & CRAVES  & CRAVES & \multicolumn{5}{c}{Ours, synt} \\
    & synt \cite{zuo2019craves} &  synt+real* \cite{zuo2019craves} & f=500  & f=1000  & f=1500  & f=2000  & \textcolor{gray}{f=best} \\
    \hline
    & 81.61 & 88.89 & 87.34 & \textbf{88.97} & 87.37 & 85.49  & \textcolor{gray}{92.91} \\
    \hline
  \end{tabular}
  \vspace{-1.0em}
  \caption{\small PCK@0.2 on the CRAVES-Youtube dataset~\cite{zuo2019craves}.}
  \label{tab:craves_youtube}
  \vspace{-1.2em}
\end{table}

We also evaluated our method on CRAVES-youtube. %
On this
dataset, the camera intrinisic parameters are unknown and cannot be used for
projecting the estimated robot pose into the 2D image. We therefore report
results for different hypothesis of (fixed) focal lengths for all the images of the
dataset, as well as using an oracle (f=best) which selects the best focal length for each image. Results are reported in Table~\ref{tab:craves_youtube}. For 2D keypoints, our method for $f=1000$ achieves results superior to CRAVES while not requiring real training images. Our method also clearly outperforms CRAVES when selecting the best focal length. 3D ground truth is not available, but similar to CRAVES-lab we could expect large improvements in 3D accuracy.

\vspace{\negative_space_paragraph}
\paragraph{Experiments on 7DoF+ robots.} We also train our method for jointly predicting 6D pose and joint angles for the
robots considered in Section~\ref{sec:exp_known_joint_angles}. We evaluate the 6D pose {\em and} joint angles accuracy using ADD. %
Results are reported in Table~\ref{tab:comparison_dream} in blue (last column). For the 7DoF robotic arms (Kuka and Panda), these results demonstrate a competitive or superior ADD accuracy compared to~\cite{lee2020camera} for inferring the 3D geometry of a known robot, but our method does not require known joint angles. %
The more complex 15 DoF Baxter robot remains challenging although our qualitative results often show reasonable alignments. {We discuss the failure modes of our approach in the appendix.}

\subsection{Analysis of parametrization choices}
 \label{sec:exp_ablations}

\begin{table}
  \small
  \begin{minipage}{.3\linewidth}
    \centering
    \begin{tabular}{lllc}
      \hline
      Reference point & ADD \\
      \hline
      on Root $\mathcal{P}_0$  & 75.02  \\
      on Middle $\mathcal{P}_4$  & 79.45 \\
      on Hand $\mathcal{P}_7$ & 00.00 \\
      \hline
      Centroid (ours) & \textbf{80.54} \\
      \hline
    \end{tabular}
    \vspace{-1em}
    \caption*{(a)}
  \end{minipage}%
  \begin{minipage}{.8\linewidth}
    \centering
    \begin{tabular}{lcc}
      \hline
      Reference point & \makecell{Volume \\ $(cm^3)$} & ADD\\
      \hline
      on $\mathcal{P}_5$ & 3092 & 74.40 \\
      on $\mathcal{P}_2$ & 2812 & 75.06 \\
      on $\mathcal{P}_1$ & 2763 & 74.89 \\
      on $\mathcal{P}_0$ & 2660 & 75.02 \\
      on $\mathcal{P}_4$ & 2198 & 79.45 \\
      \hline
      Centroid (ours)  &
      - & \textbf{80.54} \\
      \hline
    \end{tabular}
    \vspace{-0.8em}
    \caption*{(b)}
\end{minipage}
\vspace{-1.2em}
\caption{\small \textbf{Analysis of the choice of reference point $O$.} Networks are trained and evaluated with known joint angles as in Section~\ref{sec:exp_known_joint_angles}. The reference point is placed on (a) a naively chosen part and (b) on one of the 5 largest parts. Our strategy of using the centroid of the imaged robot performs the best.}
\label{tab:known_joints_ablations}
\vspace{-1.2em}
\end{table}

We analyze our method on the Panda-ORB dataset: it is the largest real dataset containing significant variations in joint angles and camera viewpoints and the Panda robot has a long kinematic chain with 8 DoF.
We study the choice of reference point $O$ for the 6D pose update and the choice of the anchor part (see Section~\ref{sec:choice_of_anchor}).

\vspace{\negative_space_paragraph}
\paragraph{Reference point.} We train different networks with the reference point
at the origin of the root $\mathcal{P}_0$, the part in the middle of the kinematic chain $\mathcal{P}_4$ and at the end of the kinematic chain $\mathcal{P}_7$.  Results are reported in Table~\ref{tab:known_joints_ablations}(a). We observe that the performance indeed depends on the choice of the  reference point and our approach of using the centroid of the robot as the reference point performs the best. The network trained with the ``Hand'' part ($\mathcal{P}_7$, the end effector) as a reference point fails to converge because this part is often difficult to identify in the training images and its pose cannot be inferred from any other part because the robot is not a rigid object.
We investigate picking the reference point on one of the five largest parts (measured by their 3D volume which is correlated with 2D visibility) in Table~\ref{tab:known_joints_ablations}(b) again demonstrating our approach of using the centroid of the robot performs better than any of these specific parts.

\begin{table}
  \small
  \begin{minipage}{.5\linewidth}
    \centering
    \begin{tabular}{lcc}
      \hline
      Anchor & \makecell{Volume \\ $(cm^3)$} & ADD\\
      \hline
      $\mathcal{P}_5$ & 3092 & 68.01 \\
      $\mathcal{P}_2$ & 2812 & 65.56 \\
      $\mathcal{P}_1$ & 2763 & 60.40 \\
      $\mathcal{P}_0$ & 2660 & 57.44 \\
      $\mathcal{P}_4$ & 2198 & \textbf{69.54}\\
      $\mathcal{P}_7$ & 637 & 63.40\\
      \hline
    \end{tabular}
    \vspace{-0.8em}
    \caption*{(a)}
  \end{minipage}%
  \begin{minipage}{.5\linewidth}
    \centering
    \begin{tabular}{lllc}
      \hline
      Anchor & ADD \\
      \hline
      Root part $\mathcal{P}_0$ & 57.44\\
      Middle part $\mathcal{P}_4$ & 69.54  \\
      Hand part $\mathcal{P}_7$ & 63.40 \\
      \hline
      Random (all) & 64.28 \\
      Random (5 largest) & 70.39 \\
      Random (3 largest) & \textbf{71.36}  \\
      \hline
    \end{tabular}
    \vspace{-1em}
    \caption*{(b)}
  \end{minipage}%
\vspace{-1.2em}
\caption{\small \textbf{Analysis of the choice of the anchor part.} Networks are trained and evaluated with unknown joint angles as in Section~\ref{sec:exp_unknown_joint_angles}. (a) Results when one fixed anchor part is used during training and testing. (b) Randomly selecting the anchor part among a given set of largest robot parts during refinement in both training and testing.
  \vspace{-2em}
}
\label{tab:anchor_selection_results}
\end{table}

\paragraph{Choice of the anchor part.}
\label{sec:unknown_joints_ablations}
\vspace{\negative_space_paragraph}
Table~\ref{tab:anchor_selection_results} reports results using different strategies for chosing the anchor part during training and testing.
First, in~\ref{tab:anchor_selection_results}(a) we show that choosing different parts as one (fixed) anchor results in significant variation in the resulting performance. 
To mitigate this issue we consider in~\ref{tab:anchor_selection_results}(b) a strategy where the anchor is picked randomly among the robot parts at each iteration (both during training and testing). This strategy performs better than always naively selecting the root $\mathcal{P}_0$ as anchor. By restricting the sampled anchors to the largest parts, our automatic strategy can also perform better than the best performing part $\mathcal{P}_4$.

\vspace{-.4em}
\section{Conclusion}

We have introduced a new {\em render \& compare} approach to estimate the joint angles and the 6D camera-to-robot pose of an articulated robot from a single image demonstrating significant improvements over prior state-of-the-art for this problem. These results open-up exciting applications in visually guided manipulation or collaborative robotics without fiducial markers or time-consuming hand-eye calibration. To stimulate these applications, we released the training code as well as the pre-trained models for commonly used robots.

\label{sec:conclusion}

{\textbf{Acknowledgments.} This work was partially supported by the HPC resources from GENCI-IDRIS (Grant 011011181R1), the European Regional Development
Fund under the project IMPACT (reg. no. CZ.02.1.01/0.0/0.0/15 003/0000468), Louis Vuitton ENS Chair on Artificial Intelligence, and the French government under management of Agence Nationale de la Recherche as part of the "Investissements d'avenir" program, reference ANR-19-P3IA-0001 (PRAIRIE 3IA Institute).}

{\small
\bibliographystyle{ieee_fullname}
\bibliography{biblio}
}

\appendix
\section*{Appendix}

This appendix is organized as follows. In
Section~\ref{sec:pose_update}, we provide the complete set of equations for the pose update and its
relation to the reference point introduced in the main paper. In Section~\ref{sec:pose_loss}, we give the
details of the pose loss $\mathcal{L}_{a}$ used in equation~\eqref{eq:loss_general} in the main paper.
Sections~\ref{sec:cropping},~\ref{sec:state_initialization},
~\ref{sec:training_strategy} and~\ref{sec:evaluation_details} give details of the cropping strategy, the state
initialization, training strategy and evaluation, respectively. In Section~\ref{sec:benefits_iterative}, we provide additional experiments showing the benefits of the iterative formulation and studying the importance of running the refiner network for multiple iterations during training.  In Section~\ref{sec:real_applications}, we discuss the applicability of our approach to real robotic problems. Section~\ref{sec:qualitative} presents additional qualitative examples randomly selected from the images in the datasets introduced in Section~\ref{sec:experiments} of the main paper. Finally, we discuss and illustrate the main failure modes of our approach in Section~\ref{sec:failure_modes}. Additional examples are in the supplementary video~\cite{projectpage}.

\section{Pose update}
\label{sec:pose_update}
Any rigid motion can be modeled by a
transformation in $SE(3)$. The 6D pose of the anchor part of the robot at iteration $k+1$ is therefore defined by the composition of its current transformation w.r.t. to the camera coordinate system $\mathcal{T}_{C,a}^k$ at iteration $k$ composed with a pose update transformation $\Delta \mathcal{T}\in SE(3)$:
\begin{equation}
  \mathcal{T}_{C,a}^{k+1} = \mathcal{T}_{C,a}^k \circ \Delta \mathcal{T},
  \label{eq:composition_two_se3}
\end{equation}
as introduced in Section~\ref{sec:framework_overview} in the main paper.
In the following we present the equations that define this pose update.
We follow the parametrization of DeepIM's \cite{li2018deepim} pose update and explicitly parametrize the pose update using a 3D reference point we call $O^k$. We choose to use the centroid of the estimated robot at the $k$-th iteration as $O^k$, but other choices are also possible.  Our neural network refiner predicts parameters $v_x$, $v_y$, $v_z$ that are used to compute the 3D translation $\Delta t$. The refiner also predicts the 3D rotation $\Delta R$ that together with $\Delta t$ define the pose update $\Delta \mathcal{T}$ in~\eqref{eq:composition_two_se3}. The parameters $v_x$ and $v_y$ correspond to the displacement in the image plane (in pixels) of the reference point $O^k$ at iteration $k$ and $v_z$ corresponds to a relative update of the depth of $O^k$, again relative to the camera coordinate system:
\begin{eqnarray}
  v_x & = &  u_{{O^k},x}^{k+1} - u_{{O^k},x}^{k} = f_x^{c}\left( \frac{x_{O^k}^{k+1}}{z_{O^k}^{k+1}} -  \frac{x_{O^k}^{k}}{z_{O^k}^{k}} \right) \\
  v_y & = & u_{{O^k},y}^{k+1} - u_{{O^k},y}^{k} = f_y^{C}\left( \frac{y_{O^k}^{k+1}}{z_{O^k}^{k+1}} -  \frac{y_{O^k}^{k}}{z_{O^k}^{k}} \right) \\
  v_z & = & \frac{z_{O^k}^{k+1}}{z_{O^k}^{k}},
\end{eqnarray}
where $u_{{O^k}}^{k}=(u_{{O^k},x}^{k},u_{{O^k},y}^{k})$ is the 2D projection onto the image plane of the 3D point $O^k$ before applying the pose update, $u_{{O^k}}^{k+1}$ the 2D reprojection onto the image plane of $O^{k}$ after applying the pose update, $\begin{bmatrix}x_{O^k}^{k},y_{O^k}^{k},z_{O^k}^{k}\end{bmatrix}$ are the 3D coordinates of the reference point $O^{k}$ before the pose update expressed in the camera frame, and $\begin{bmatrix}x_{O^k}^{k+1},y_{O^k}^{k+1},z_{O^k}^{k+1}\end{bmatrix}$ are the 3D coordinates of the reference point $O^{k}$ after the pose update expressed in the camera frame. {$f_x^{C}$ and $f_y^{C}$ are the intrinsic parameters of the virtual cropped camera that are assumed known and fixed.}

From these equations, we can derive the update of the 3D translation of the 3D point $O^k$:
\begin{equation}
  t_{O^k}^{k+1} = t_{O^k}^{k} + \Delta t_{O^k},
\end{equation}
with the $x$, $y$ and $z$ components of translation update $ \Delta t_{O^k}$  defined as :
\begin{alignat}{2}
  \Delta t_{{O^k},x} & = & x^{k+1}_{O^k} - x^{k}_{O^k}   & =  \frac{1}{f_x^{C}} v_x v_z z^{k}_{O^k} + x^{k}_{O^k} (v_z - 1) \label{eq:delta_x} \\
  \Delta t_{{O^k},y} & = & y^{k+1}_{O^k} - y^{k}_{O^k}   & =  \frac{1}{f_y^{C}} v_y v_z z^{k}_{O^k} + y^{k}_{O^k} (v_z - 1) \label{eq:delta_y}\\
  \Delta t_{{O^k},z} & = & z^{k+1}_{O^k} - z^k_{O^k}     & =  z^{k}_{O^k} (v_z - 1). \label{eq:delta_z}
\end{alignat}
Note that $\Delta t_{O^k}$ depends on the 3D point $O^{k}$. Hence the network has to learn to internally infer this point to predict consistent transformation updates. 
The rotation update $\Delta R$ (parametrized by three angles) predicted by the refiner network is applied around the 3D reference point $O^k$. This defines the transformation of any 3D point on the anchor. Let's consider a point $a$ on the anchor part in position $t^k_a$ at iteration $k$, its position at iteration $k+1$ is given by:
\begin{eqnarray}
t_{a}^{k+1} =  \Delta R (t_a^{k} - t_{O^k}^{k}) + t_{O^k}^{k} + \Delta t_{O^k}, \label{eq:delta_t}
\end{eqnarray}
where $\Delta R$ is the rotation matrix predicted by the network using the same rotation parametrization used in~\cite{labbe2020cosypose,Zhou2018-eg}. The equation~\eqref{eq:delta_t} can also be used to define the rotation matrix of the anchor part with respect to the camera:
\begin{eqnarray}
R_{C,a}^{k+1}  = \Delta R R_{C,a}^{k}, \label{eq:delta_R}
\end{eqnarray} where $R_{C,a}^{k}$ defines the rotation of the anchor part with respect to the camera before the pose update, $R_{C,a}^{k+1}$ defines the rotation of the anchor part with respect to the camera after the pose update and $\Delta R$ is the rotation matrix predicted by the network.
The equations ~\eqref{eq:delta_t} and \eqref{eq:delta_R} fully define the pose update given by equation~\eqref{eq:composition_two_se3} as a function of the network predictions $v_x$, $v_y$, $v_z$ and $\Delta R$, and the reference point $O^{k}$. For computing the loss in the next section, we write the pose $\mathcal{T}_{C,a}^{k+1}$ of the anchor part $a$ in the camera coordinate system $C$ at iteration $k+1$ (after the pose update) as
\begin{eqnarray}
  \mathcal{T}_{C,a}^{k+1} = \mathcal{U}([v_x, v_y, v_z], \Delta R),
  \label{eq:pose_update}
\end{eqnarray}
where $\mathcal{U}$ expresses the updated pose of the anchor as a function of all network predictions $v_x$, $v_y$, $v_z$, $\Delta R$, and can be computed in a closed form using the equations derived above.

\section{Pose loss}
\label{sec:pose_loss}
We define the following distance $D_a(\mathcal{T}_1, \mathcal{T}_2)$ to measure the distance between two transformations $\mathcal{T}_1$ and $\mathcal{T}_2$ using the 3D points $\mathcal{X}_a$ on the anchor part $a$ :
\begin{eqnarray}
  D_a(\mathcal{T}_1, \mathcal{T}_2) = \frac{1}{|\mathcal{X}_a|} \sum_{x\in \mathcal{X}_a}|\mathcal{T}_1 x - \mathcal{T}_2 x |,
\end{eqnarray}
where $|\cdot|$ is the $L_1$ norm.
The pose loss $\mathcal{L}_{a}$ in equation (3) in the main paper follows~\cite{labbe2020cosypose} and is written as:
\begin{align}
  \mathcal{L}_{pose} & = D_a(\mathcal{U}([v_x,v_y,\hat{v}_z], \Delta \hat{R}), \hat{T}_{C,a}) \label{eq:loss-xy}\\
                    & + D_a(\mathcal{U}([\hat{v}_x,\hat{v}_y,v_z], \Delta \hat{R}), \hat{T}_{C,a})\label{eq:loss-z}  \\
                    & + D_a(\mathcal{U}([\hat{v}_x,\hat{v}_y,\hat{v}_z], \Delta R), \hat{T}_{C,a})\label{eq:loss-R},
\end{align}
where $\mathcal{U}$ defines the pose update as explained in~\eqref{eq:pose_update} in Section~\ref{sec:pose_update}, $\hat{T}_{C,A}$ is the ground truth 6D pose of the anchor with repect to the camera, $v_x, v_y, v_z, \Delta R$ are the parameters of the pose update predicted by the network, and $\hat{v}_x, \hat{v}_y, \hat{v}_z, \Delta \hat{R}$ are the values of these updates that would lead to the ground truth pose. The different terms of this loss separate (or disentangle) the influence of different subsets of parameters on the loss. In detail, the first term ~\eqref{eq:loss-xy} of the loss considers only the $xy$ translation of the anchor part, where the rest of the parameters are fixed to their ground truth values, $\hat{v}_z$ and $\Delta \hat{R}$ . The second term ~\eqref{eq:loss-z} of the loss considers only the relative depth of the anchor part where the rest of the parameters are fixed to their ground truth values. Finally, the last term~\eqref{eq:loss-R} of the loss considers only the rotation update $\Delta R$ \eqref{eq:loss-R} where the rest of the parameters are fixed to their ground truth values.

\section{Cropping strategy}
\label{sec:cropping}
\begin{figure*}[t]
  \centering
  \includegraphics[width=1.0\linewidth]{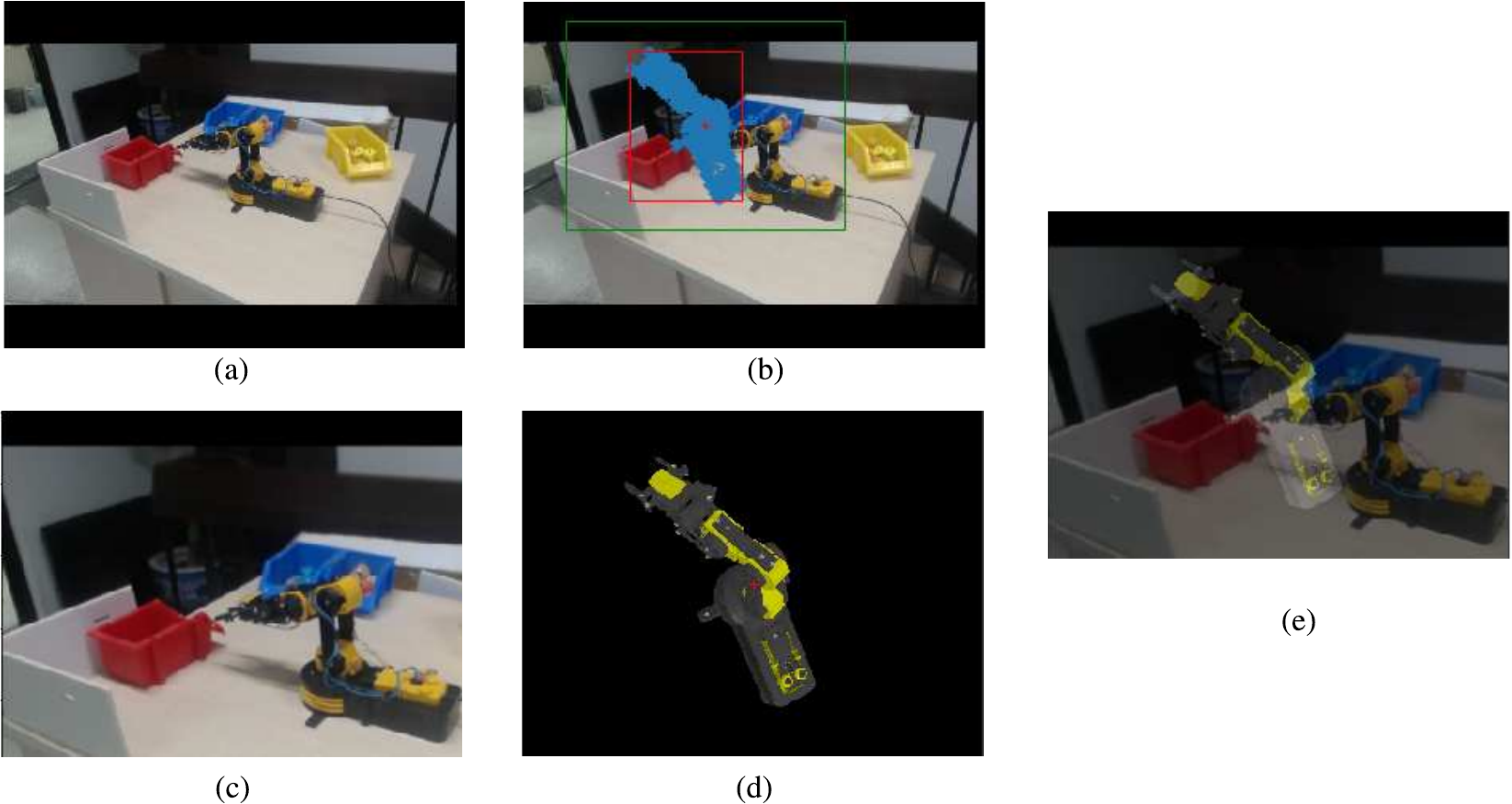}
  \caption{\small Cropping strategy. (a) input RGB image, (b) illustration of the cropping strategy: bounding box $[u_{1},u_{2}]$ (shown in red) defined by the reprojection of the robot in the input state to the image (blue points). Bounding box defining the crop is shown in green. (c) The cropped input image, (d) Cropped rendered image, (e) The crop and the input image are overlaid. These two images are given as input to our network (the anchor mask is not displayed).}
  \label{fig:cropping}
\end{figure*}

Let $u_{O} = (u_{O,x}, u_{O,y})$ be the 2D projection of the 3D centroid $O$ of the robot by the camera with intrinsics $K$. We use a cropping strategy similar to DeepIM but using only the projection of the 3D points of the model in the current pose estimate to define the bounding box of the crop. Let $u_1= (u_{1,x}, u_{1,y})$, $u_2= (u_{2,x}, u_{2,y})$ be the coordinates of the upper left and lower right corners of the bounding box defined by the robot projection in the image, respectively. We define:

\begin{eqnarray}
  \Delta u_x & = & max(u_{dist,x}, u_{dist,y}/r) \cdot 2 \lambda, \\
  \Delta u_y & = & max(u_{dist,x}/r, u_{dist,y}) \cdot 2 \lambda,
\end{eqnarray}
where
\begin{eqnarray}
  u_{dist,x} & = & max(|u_{1,x} - u_{O,x}|, |u_{2, x} - u_{O,x}|), \\
  u_{dist,y} & = & max(|u_{1,y} - u_{O,y}|, |u_{2, y} - u_{O,y}|),
\end{eqnarray}
$r=4/3$ is the aspect ratio of the crop and $\lambda=1.4$ following
DeepIM~\cite{li2018deepim}. The crop is centered at $u_O$, of width $\Delta u_x$ and height $\Delta u_y$. An example of a crop is given in Figure~\ref{fig:cropping}.

\section{State initialization}
\label{sec:state_initialization}
\begin{figure*}[t]
  \centering
  \includegraphics[width=1.0\linewidth]{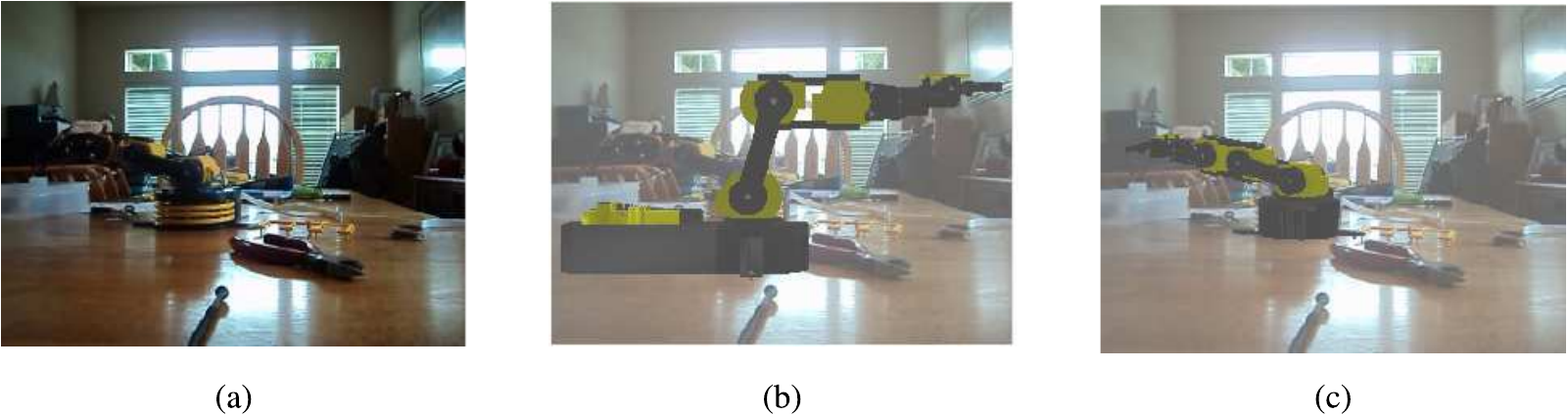}
  \caption{\small Example of initialization. (a) Input RGB image, (b) robot in the initial state, (c) output of RoboPose (pose and joint angles are predicted) after $K=10$ refinement iterations.}
  \label{fig:initialization}
\end{figure*}

Let $u_{det}=(u_{{det},x},u_{{det},y})$ and $(\Delta u_{{det}}=\Delta u_{{det},x},\Delta u_{{det},y})$ define the center and the size of the approximate 2D bounding box of the robot in the image. In our experiments, there is only one robot per image and thus we use the entire image as the 2D bounding box and denote the bounding box $u_{det}$.
The orientation of the base of the robot $\mathcal{P}_0$ is set parallel to the axes of the camera with the $z$ axis pointing upwards. %
The centroid of the robot $O$ is set to match the center of the bounding box $u_{det}$. We make the first hypothesis of the depth of the centroid by setting $z_{C,O}^{guess}=1 \mathrm{m}$ and use this initial value to estimate the coordinates $x$ and $y$ of the centroid in the camera frame:
\begin{eqnarray}
  x_{C,O}^{guess} =  u_{{det},x} \frac{z_{C,O}^{guess}}{f_x} \label{eq:init_x}\\
  y_{C,O}^{guess} =  u_{{det},y} \frac{z_{C,O}^{guess}}{f_y} \label{eq:init_y}.
\end{eqnarray}
We then update the depth estimate $z_{C,O}^{guess}$ using the following simple strategy. We project the points of the robot using the initial guess we have just defined. These points define a bounding box with dimensions $\Delta u_{guess,x}=(\Delta u_{guess,x}, \Delta u_{guess,y})$ and the center remains unchanged $u_{guess}=u_{det}$ by construction. We compute an updated depth of the centroid such that its width and height approximately match the size of the 2D detection:
\begin{eqnarray}
  z_{C,M}^0=  z_{C,O}^{guess} \frac{1}{2} \left( f_x \frac{\Delta u_{guess,x}}{\Delta u_{det,x}} + f_y \frac{\Delta u_{guess,y}}{\Delta u_{det,y}}  \right)
\end{eqnarray}
and use this new depth to compute $x_{C,O}^0$ and $y_{C,O}^0$ using equations~\eqref{eq:init_x} and~\eqref{eq:init_y} that were used to define $ x_{C,O}^{guess}$ and $ y_{C,O}^{guess}$.  The initial state $\mathcal{S}^0$ of the robot is defined by $(x_{C,M}^0, y_{C,M}^0, z_{C,M}^0)$, $R_{C,O}^0=Id$, and the initial joint angles are set (when they are not measured) to $q^0=\frac{q^+ + q^-}{2}$, where $q^+$ and $q^-$ define the interval of the robot articulation angles. An example of initialization is shown in Figure~\ref{fig:initialization}.

{In the experiments of Section~\ref{sec:experiments} of the main paper, we use the entire image as the 2D bounding box because the test images show a single robot. Please note however that it would be straightforward to process multiple bounding boxes like~\cite{labbe2020cosypose} for multiple rigid objects. If multiple robots are in the input bounding box (e.g. as in some of the examples in the results video on the project webpage~\cite{projectpage}), our iterative refinement typically converges to the largest robot in the image.}

\section{Training strategy}
\label{sec:training_strategy}
The training of the neural networks is performed with Adam~\cite{Kingma2014-xz} using a learning rate of $0.003$ and batches of 1408 images, split on 44 GPUs. The networks are trained for 60k iterations. Following recommendations from~\cite{Goyal2017-of} for distributed training, we use a warm up-phase: the learning rate is linearly increased from to $0.$ to $0.003$ during the first 5k iterations. The learning rate is reduced to 0.0003 at 45k iterations. In order to speed up the training, we start by training with $K=1$ refinement iterations and add other iterations $K=2$, $K=3$ at 15k and 30k iterations respectively. We found that starting to train with $K=1$ and increasing the number of refinement iterations during training has no effect on the results, but allows to train the refiner network faster.

\section{Evaluation details}
\label{sec:evaluation_details}
Next, we give more details about the evaluation metric and the evaluation protocol of our comparison with DREAM~\cite{lee2020camera} and CRAVES~\cite{zuo2019craves}.

\subsection{Comparison with DREAM} The average distance (ADD) metric measures the distance between 3D keypoints on the robot transformed with the predicted and ground truth 6D poses. The 3D keypoints are the same as in~\cite{lee2020camera} for the Kuka, Panda and Baxter robots and are defined at the locations of robot joints. The ADD errors of the 3D keypoints for the robot depicted in the image are averaged, and we report the area-under-the-curve of the ADD pass rate vs. threshold following~\cite{lee2020camera}. We refer to \cite{lee2020camera,Xiang2018-dv} for more details about this metric. A slight difference with the evaluation protocol of~\cite{lee2020camera} is that all images are considered as possible evaluation targets, whereas~\cite{lee2020camera} discards images where there are fewer than 4 robot keypoints visible in the ground truth because PnP cannot be solved in this situation. This favors methods that rely on keypoints and does not fully consider the difficulty of estimating the pose of robots in situations with high self-occlusions, a situation that our method can handle. We thus evaluate ADD on all images, even those with fewer than four visible robot keypoints. The models from DREAM~\cite{lee2020camera} were re-evaluated with this protocol and in practice the difference is only minor compared to the results reported in DREAM~\cite{lee2020camera}, less than $2\%$ for all datasets. For reproducing DREAM results reported in Table 1 of the main paper, we used the provided code and pre-trained models\footnote{https://github.com/NVlabs/DREAM} using the instructions to reproduce the results from the paper, e.g. using the ``shrink-and-crop'' cropping strategy. Note that while evaluating these models, we found some results to be higher than reported in the original paper, e.g. the ResNet101-H is the best model on the largest real dataset Panda-ORB (69.05) but only the VGG19-F (61.93) was reported in~\cite{lee2020camera}. We reported results for all models of DREAM~\cite{lee2020camera} in Table 1 of the main paper.

When evaluating our method that predicts joint angles, ADD is computed using 3D keypoints on the robot that are computed using both the estimated 6D pose and the predicted joint angles using the robot forward kinematics. This provides a principled way to measure the accuracy of the 3D reconstruction of the robot without giving arbitrary weights to rotation, translation or joint angle errors.

\subsection{Comparison with CRAVES}
For the comparison with CRAVES~\cite{zuo2019craves}, we based our evaluation on the code provided with the paper\footnote{https://github.com/zuoym15/craves.ai}. The 3D metrics reported in Table 3 for CRAVES-lab are:
\begin{itemize} 
  \item Joints (degrees): Errors of the joint angle predictions over the 4 articulations $\frac{1}{4}\sum_{i=1}^{4}|q^{pred}_i - q^{gt}_i|$, where $q^{pred}_i$ is the predicted value of the joint angle for joint $i$ and $q^{gt}_i$ is the ground truth value of the joint angle for joint $i$.
  \item Trans xyz. (cm): Translation error for the base averaged over the $x,y,z$ axes: $\frac{1}{3} (|x^{pred} - x^{gt}| + |y^{pred}-y^{gt}| + |z^{pred} - z^{gt}|)$, where "pred" denotes the predicted and "gt" the ground truth values. 
  \item Trans norm. (cm): Translation error for the base computed using the $L_2$ norm: $||t^{pred} - t^{gt}||_2$.
  \item Rot. (degrees): Errors between the Euler angles that define the rotation of the base with respect to the camera $\frac{1}{3}(|\theta_x^{pred} - \theta_x^{gt}| + |\theta_y^{pred} - \theta_y^{gt}| + |\theta_z^{pred} - \theta_z^{gt}|)$, where again "pred" denotes the predicted and "gt" the ground truth values. 
\end{itemize}

We found that the "Joints'', ``Trans xyz.'' and "Rot." metrics reported in~\cite{zuo2019craves} under "3D pose estimation errors" were an average computed only over the top 50\% images with the best predictions according to the joint angle errors. We also report the errors averaged over {\em all} images of the dataset (all) in Table 3 of the main paper as our method can successfully recover the 3D configuration of the robot in all images of the dataset because it does not rely on solving the 2D-to-3D nonlinear, nonconvex optimization problem that is difficult to solve in some situations.

On CRAVES-Youtube only the visible 2D keypoints are considered as estimation targets to provide a fair comparison with ``Youtube-vis'' of~\cite{zuo2019craves}. The results that we reproduced with the provided pre-trained models were slightly lower (around 0.5\% lower) than reported in the paper for the PCK@0.2 metric, but we kept the (higher) results reported in the paper when comparing with their method.

{
\section{Benefits of the iterative formulation}
\label{sec:benefits_iterative}

In this section, we investigate the benefits of the number of refiner iterations at training and test time. In the following, we denote as $K_{train}$ the number of refiner network iterations used during training (denoted $K$ in equation~\eqref{eq:loss_general} of the main paper) and as $K_{test}$ the number of iterations used at test time. We experimentally evaluate the influence of $K_{train}$ and $K_{test}$ using the ADD metric on the Panda-ORB dataset with unknown joint angles, similar to the experimental set-up reported in Section~\ref{sec:exp_unknown_joint_angles}. We trained four networks with $K_{train}=1/2/3/5$ and report the results for a varying number of iterations at test time for each network. Results are reported in Table~\ref{tab:benefits_iterative_formulation}.

First, we observe that for each refiner network (each row in the table), the accuracy is significantly improved by using multiple iterations at test time as opposed to running the network for a single iteration ($K_{test}=1$) in a single-shot regression fashion. These results demonstrate that the iterative formulation is crucial to the success of the approach.

Second, we observe that running the network for multiple iterations during training significantly improves the results. For example, for $K_{test}=10$, the ADD results are improved from $41.25$ for $K_{train}=1$ to $70.77$ for $K_{train}=5$. Using multiple training iterations without backpropagating through the renderer augments the distribution of errors between the input state and the ground truth state with actual errors the refiner makes. These results show the importance of training the refiner network to correct both the large errors (between the initial state $\mathcal{S}^{0}$ and the ground truth state $\mathcal{S}^{gt}$) as well as smaller errors (between the prediction at iteration k, $\mathcal{S}^{k}$, and $\mathcal{S}^{gt}$) to simulate what the network is going to see at test-time where the network is run for multiple iterations. In our experiments in the paper, we used $K_{train}=3$ as it is a good trade-off between performance and training speed.

\setlength{\tabcolsep}{4pt}
\begin{table}[t]
  \small
  \centering
  \begin{tabular}{c|ccccc}
    \hline
    {\footnotesize \backslashbox{$K_{train}$}{$K_{test}$}} & 1  & 2 & 3 & 5 & 10 \\
    \hline
    \hline
    1 &  0.43        &  24.31        & 34.72     & 39.77 & 41.25 \\
    \hline
    2 &  1.70        &  32.22        & 52.08     & 62.76 & 65.73 \\
    \hline
    3 &  1.14        &  25.51        & 49.22     & 65.64 & 70.39 \\
    \hline
    5 &  0.61        &  15.80        & 36.81     & 61.08 & {\bf 70.77} \\
    \hline
  \end{tabular}
  \vspace{-1.0em}
  \caption{\small \textbf{Benefits of the iterative refinement.} We study the influence of the number of training  and testing iterations, denoted $K_{train}$ and $K_{test}$, respectively. We report the $ADD$ metric (higher is better) on the Panda ORB dataset with unknown joint angles. The best performing set-up is shown in {\bf bold}.}
  \label{tab:benefits_iterative_formulation}
  \vspace{-1.0em}
\end{table}

}

{
\section{Applications in robotic set-ups}
  \label{sec:real_applications}
  Similar to related work~\cite{lee2020camera,zuo2019craves}, our approach can be applied in real robotic set-ups.
  For example,  
  the offline hand-eye calibration approach DREAM~\cite{lee2020camera} has been used in~\cite{tremblay2020indirect} for a manipulation task on a real robot, and CRAVES~\cite{zuo2019craves} demonstrates visually guided closed-loop control of a low-cost robot without reliable joint angle measurements. Our approach significantly improves over the accuracy of DREAM~\cite{zuo2019craves} and CRAVES~\cite{lee2020camera}, and therefore we expect to also improve the capabilities of robotic systems similar to~\cite{tremblay2020indirect,zuo2019craves}. Please note that while our paper focuses on the robot state estimation from a single image, it is also suited for real-time and online applications as explained in Section~\ref{sec:exp_known_joint_angles}. In this scenario (also illustrated in the video on the project webpage~\cite{projectpage}), our predictions could be temporally smoothed to reduce the jitter in the predictions by applying a simple temporal low-pass filter similar to~\cite{tremblay2020indirect}.
}

\section{Qualitative examples}
\label{sec:qualitative}

\begin{figure*}[t]
  \centering
  \includegraphics[width=1.0\linewidth]{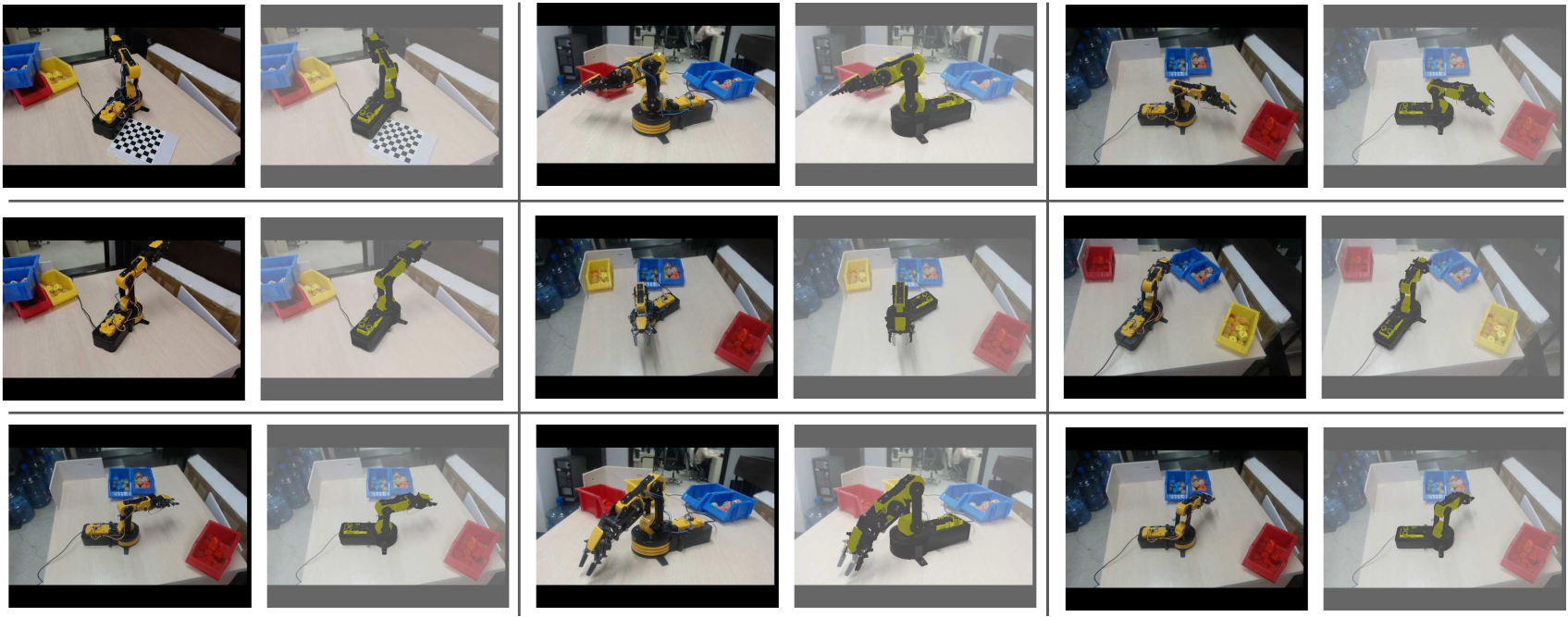}
  \caption{\small {\bf Random selection of examples on the CRAVES-lab datasets}. Both joint angles and the 6D pose of the robot are predicted from the input image. For each of the 9 examples, we show the input RGB image (left) and the predicted state of the robot within the 3D scene (right). We illustrate the predicted state (right) by overlaying the articulated CAD model of the robot in the predicted state over the input image.}
  \label{fig:examples_craves_lab}
\end{figure*}

\begin{figure*}[t]
  \centering
  \includegraphics[width=1.0\linewidth]{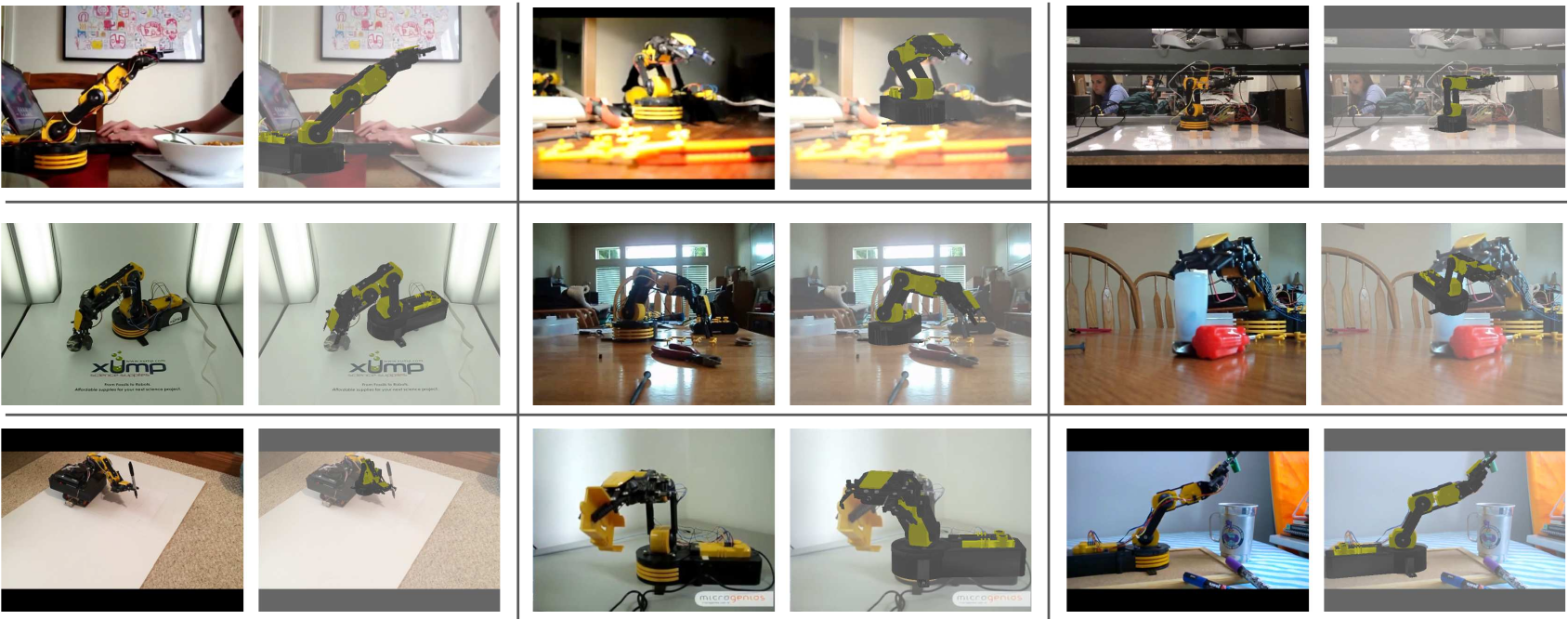}
  \vspace{-2em}
  \caption{\small {\bf Random selection of examples on the CRAVES-youtube dataset}. Both joint angles and the 6D pose of the robot are predicted from the input image. For each of the 9 examples, we show the input RGB image (left) and the predicted state of the robot within the 3D scene (right). We illustrate the predicted state (right) by overlaying the articulated CAD model of the robot in the predicted state over the input image.}
  \label{fig:examples_craves_youtube}
\end{figure*}

\begin{figure*}[t]
  \centering
  \includegraphics[width=1.0\linewidth]{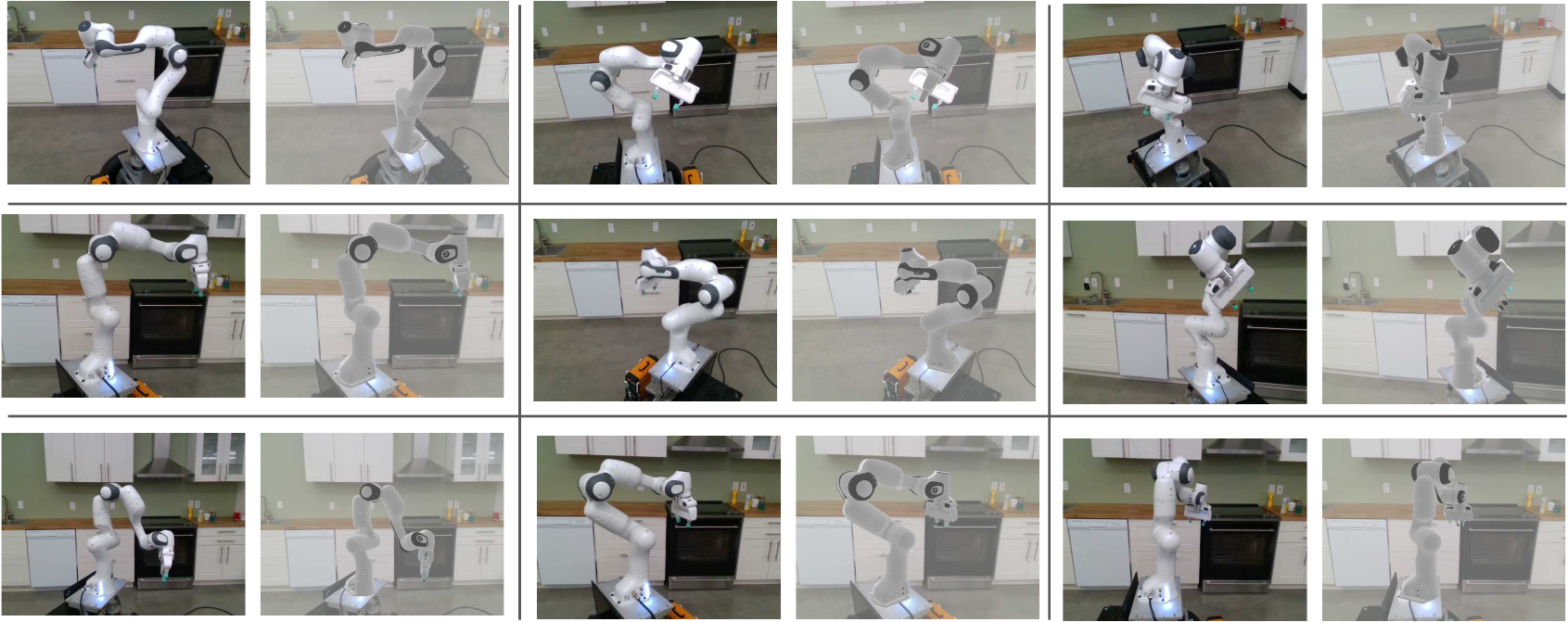}
  \caption{\small {\bf Random selection of examples on the Panda ORB dataset}.  Both joint angles and the 6D pose of the robot are predicted from the input image. For each of the 9 examples, we show the input RGB image (left) and the predicted state of the robot within the 3D scene (right). We illustrate the predicted state (right) by overlaying the articulated CAD model of the robot in the predicted state over the input image.}
  \label{fig:examples_panda_orb}
\end{figure*}

\begin{figure*}[t]
  \centering
  \includegraphics[width=1.0\linewidth]{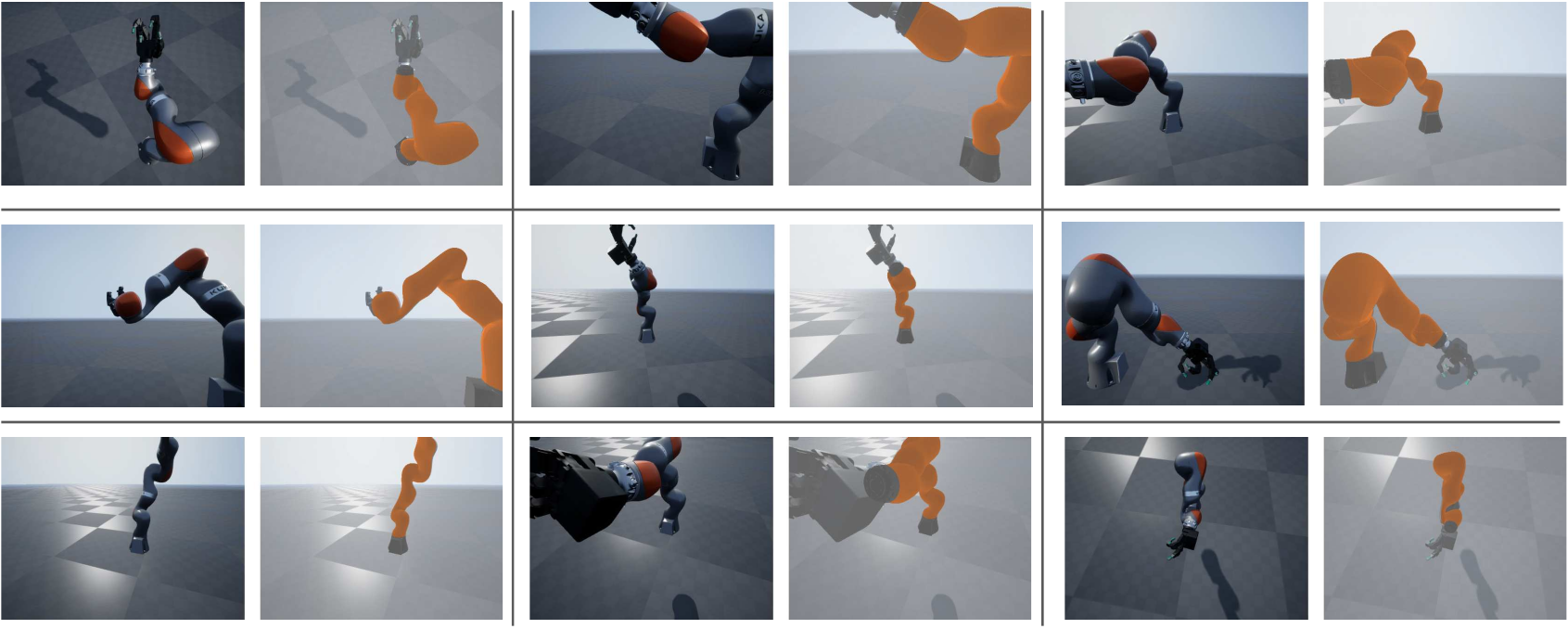}
  \caption{\small {\bf Random selection of examples on the Kuka Photo dataset}.  Both joint angles and the 6D pose of the robot are predicted from the input image. For each of the 9 examples, we show the input RGB image (on the left) and the predicted state of the robot within the 3D scene (on the right). We illustrate the predicted state (on the right) by overlaying the articulated CAD model of the robot in the predicted state over the input image.}
  \label{fig:examples_kuka_photo}
\end{figure*}

\begin{figure*}[t]
  \centering
  \includegraphics[width=1.0\linewidth]{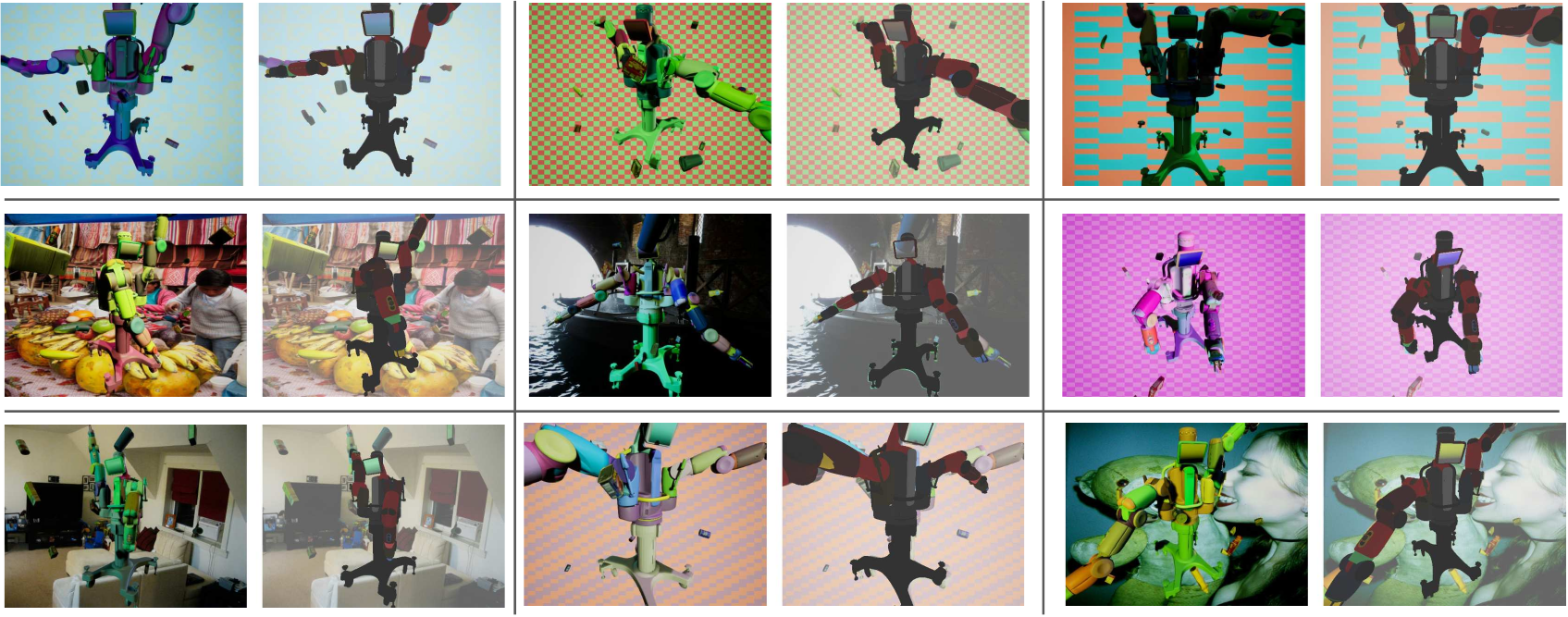}
  \caption{\small {\bf Random selection of examples on the Baxter DR dataset}.  Both joint angles and the 6D pose of the robot are predicted from the input image. For each of the 9 examples, we show the input RGB image (left) and the predicted state of the robot within the 3D scene (right). We illustrate the predicted state (right) by overlaying the articulated CAD model of the robot in the predicted state over the input image.}
  \label{fig:examples_baxter_dr}
\end{figure*}

\paragraph{Random examples.} Here we provide more qualitative examples on the datasets where we have shown the quantitative evaluation. All the presented results consider the most challenging scenario where {\em the joint angles are unknown and are predicted} together with the robot's 6D pose. For each dataset, we have randomly sampled 9 images without any further manual selection and show the input RGB image and the output state predicted by our method. Results are presented for the CRAVES-lab (Figure~\ref{fig:examples_craves_lab}), CRAVES-youtube (Figure~\ref{fig:examples_craves_youtube}), Panda ORB (Figure~\ref{fig:examples_panda_orb}), Kuka Photo (Figure~\ref{fig:examples_kuka_photo}) and Baxter DR (Figure~\ref{fig:examples_baxter_dr}) datasets.

{
\section{Failure modes}
\label{sec:failure_modes}
There are four main failure modes and limitations of our approach in the most challenging set-up with unknown joint angles, illustrated in Figure~\ref{fig:failure_modes}. First, while our method is tolerant to some amount of self-occlusion, it can fail in situations where multiple parts of the robot are occluded by external objects. Second, in some in-the-wild images (where camera intrinsics are also unknown), our iterative alignment can get stuck in a local minima. This could be improved by (i) using a better initialization (e.g. by using a separate external coarse estimation method) or by (ii) trying multiple initializations and selecting the best result. The third failure mode is related to the symmetry of individual robot parts or of entire robot configurations that are not taken into account in our method. Finally, robots with many degrees of freedom such as the 15 DoF Baxter remain challenging (please see the quantitative evaluation in  Section~\ref{sec:exp_unknown_joint_angles} in the main paper), although our qualitative results often show reasonable alignment, as illustrated in Figure~\ref{fig:examples_baxter_dr}.

\begin{figure}[t]
  \centering
  \includegraphics[width=1.0\columnwidth]{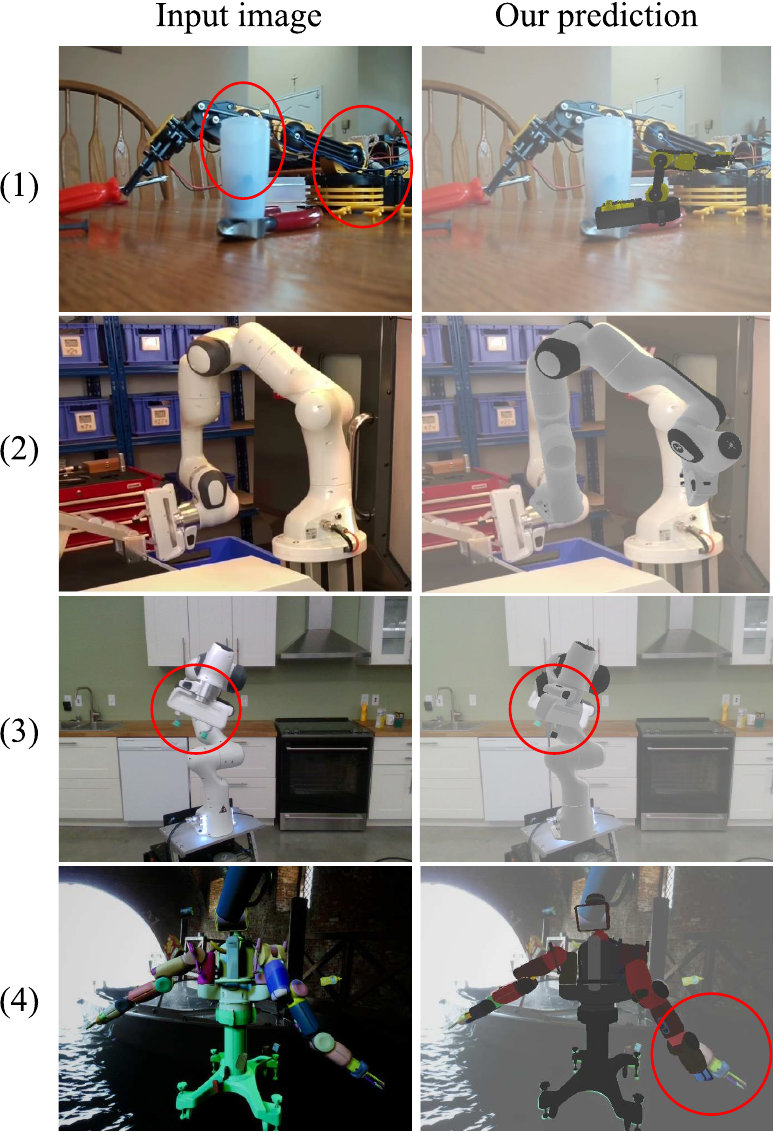}
  \caption{\small \textbf{Main failure modes.} We illustrate the four main failure modes of our approach. In (1), the severe occlusions (circled in red) of the base of the robot and of one of the parts in the middle of the kinematic chain lead to an incorrect prediction. In (2), the state of the robot is incorrectly estimated because our iterative procedure gets stuck in a local minimum. In (3), the gripper of the robot (circled in red) is a symmetric part, which leads to an incorrect estimate of the joint angle between the gripper and the rest of the robot. In (4), we illustrate an example of incorrect alignment of parts (circled in red) for the complex 15 DoF Baxter robot.}
  \label{fig:failure_modes}
  \vspace{-1.2em}
\end{figure}
}

\end{document}